\documentclass[11pt]{academic_template}
\usepackage[]{natbib}

\usepackage[utf8]{inputenc} 
\usepackage[T1]{fontenc}    
\usepackage{hyperref}       
\usepackage{url}            
\usepackage{booktabs}       
\usepackage{amsfonts}       
\usepackage{nicefrac}       
\usepackage{microtype}      
\usepackage{enumitem} 
\usepackage{amsmath}
\usepackage{graphicx}
\usepackage{algorithm}
\usepackage{nicematrix}
\usepackage{multirow}
\usepackage{algorithm}
\usepackage{algpseudocode}
\usepackage{wrapfig}
\usepackage{listings}

\newif\ifreviewchanges
\reviewchangestrue
\ifreviewchanges\fi
\DeclareRobustCommand{\revdel}[1]{%
  \ifreviewchanges
    \ifmmode\text{\textcolor{gatingblue}{\sout{\ensuremath{#1}}}}\mkern4mu\else\textcolor{gatingblue}{\sout{#1}}\fi
  \fi}
\DeclareRobustCommand{\revadd}[1]{%
  \ifreviewchanges
    \ifmmode{\color{gatingblue}\boldsymbol{#1}}\else{\color{gatingblue}\bfseries\boldmath #1}\fi
  \else #1\fi}
\usepackage{listingsutf8}
\usepackage{xcolor}

\definecolor{gatingblue}{HTML}{2056A0}

\lstdefinelanguage{json}{
    basicstyle=\ttfamily\small,
    showstringspaces=false,
    breaklines=true,
    string=[s]{"}{"},
    comment=[l]{//},
    morecomment=[s]{/*}{*/},
    keywords={true,false,null}
}

\newtcolorbox{jsonbox}{
    breakable,
    boxrule=0.4pt,
    colback=white,
    colframe=black,
    arc=0pt,
    boxsep=4pt,
    left=4pt, right=4pt, top=4pt, bottom=4pt,
    fontupper=\ttfamily\small
}

\newtcolorbox{qwendemobox}{
    boxrule=0.4pt, colback=white, colframe=gatingblue, arc=0pt,
    boxsep=2pt, left=3pt, right=3pt, top=2pt, bottom=2pt,
    before skip=4pt, after skip=4pt,
    fontupper=\ttfamily\footnotesize\color{gatingblue}
}

\title{Reward as an Agent for Embodied World Models}

\author[*]{Pu Li}
\author[*]{Zhigang Lin}
\author{Qiang Wu}
\author{Yongxuan Lv}
\author{Fei Wang}
\author[\dagger]{Shan You}

\affiliation[]{Kairos Team}

\contribution[*]{Equal contribution}
\contribution[\dagger]{Corresponding author}

\abstract{
RL for world models largely relies on conservative rollouts near the training distribution, limiting exploration and behavioral diversity. In this work, we challenge this conservative paradigm. We argue that the core limitation is not exploration itself, but the lack of reliable verification strategies to support broader exploration. Without reliable verification, policies can exploit proxy rewards through convincing videos that lack evidence of genuine task progress. On the verification side, we introduce \textbf{Reward as an Agent}, which centers evaluation on evidence-grounded task verification. It separates observed behavior from fixed task requirements, actively inspects disputed evidence, and revises judgments through tool-assisted reflection and requirement checks. Reward gates preserve supported progress while withholding credit for failure or unverifiable essential content. The framework is extensible: additional reward strategies can be integrated as tools to enrich verification and support further performance gains. On the exploration side, we introduce \textbf{Dynamic-Aware Rollout Diversification} through \textbf{DynDiff-GRPO}, which allocates stochastic exploration to dynamically salient regions to diversify trajectories. Together, these designs couple broader exploration with evidence-based reward evaluation. Training experiments demonstrate benchmark improvements across three open-source world models, while complementary offline evaluations show that Reward as an Agent improves reward reasonableness and reduces erroneous positive rewards on failure and unobservable clips.
\vspace{10mm}
}

\date{October 12, 2026}
\checkdata[Code]{\url{https://github.com/SpursLipu/Reward-as-An-Agent-for-Embodied-World-Models}}

\colorlet{themecolor}{black}
\colorlet{themeblue}{black}
\colorlet{gatingblue}{black}
\hypersetup{colorlinks=true,allcolors=black}
\AtBeginDocument{\color{black}\hypersetup{allcolors=black}}

\begin{document}

\maketitle
\newpage

\begin{spacing}{0.9}
\tableofcontents
\end{spacing}

\newpage
\begingroup\color{black}\colorlet{themecolor}{black}\hypersetup{linkcolor=black,citecolor=black,urlcolor=black}
\section{Introduction}
World models support learning, planning, and generation by predicting how observations evolve under actions. As learned simulators, they enable policy optimization and interaction learning from large-scale data without explicit physical modeling~\cite{yang2023unisim,assran2025vjepa2selfsupervisedvideo}. World-action models (WAMs) further couple visual prediction with action modeling~\cite{cen2025worldvla,ye2026dreamzero,li2026lingbotva,kim2026cosmospolicy,kairosteam2026kairosnativeworldmodel}, making reliable evaluation of predicted interactions and task outcomes essential for action learning.

Reinforcement learning (RL) offers a natural approach to world-model post-training. Group-based methods such as GRPO~\cite{shao2024deepseekmathpushinglimitsmathematical} have achieved substantial gains in language and vision-language models~\cite{Guo_2025,bai2025qwen3vltechnicalreport}, supported by reliable rewards from code execution or rule-based checks. Recent work extends these methods to visual generation~\cite{liu2025flowgrpotrainingflowmatching,xue2025dancegrpounleashinggrpovisual,li2026mixgrpounlockingflowbasedgrpo,wang2026prefgrpopairwisepreferencerewardbased,zheng2026sagegrpo,unifiedreward-flex}, but reliable reward evaluation remains a bottleneck for embodied world models.

Generated videos lack a direct correctness test: plausible appearance can conceal missing interactions, limited task progress, or physical defects. Broader RL exploration can amplify these ambiguities, allowing policies to exploit visual shortcuts that increase proxy rewards without improving behavior. Reliable evaluation therefore requires both evidence of what occurred and explicit criteria for what the task required.

Our core insight is that \textbf{exploration and verification must scale in tandem}. Reward as an Agent first records visible behavior without exposing the observer to the task instruction, while separately extracting and freezing task requirements. It then jointly assesses task completion, physical plausibility, and visual quality. To verify its judgments, the agent revisits selected frames and regions, consults CoTracker3 motion evidence and WMReward prediction-surprise feedback, and checks for omitted requirements and contradictions. Tools inform verification and reflection rather than directly contributing to the reward. The framework also supports additional tool adapters, including human-feedback reward models~\cite{he2024videoscore,liu2025improvingvideogenerationhuman}, reasoning-based evaluators~\cite{he2025videoscore2,unifiedreward-flex}, and world-model physical priors~\cite{yuan2026wmreward}, whose feedback can be checked against visual evidence.

Fixed score aggregation and explicit gates preserve useful partial progress while withholding credit for task failure or unverifiable essential content. Execution and tool faults instead produce ineligible results. This makes reward decisions inspectable, while retaining dependence on the underlying models and tools. Complementing verification, DynDiff-GRPO broadens trajectory diversity by allocating stochastic exploration to dynamically salient regions.

Our contributions are threefold:
\begin{itemize}[leftmargin=1.5em, itemsep=0.2em]
    \item \textbf{Reward-as-an-Agent Framework:} We combine task-blind observations and fixed task requirements with joint assessment, adaptive inspection, tool-assisted reflection, and scope repair. Explicit gates distinguish supported progress, zero-credit outcomes, and evaluator faults.
    \item \textbf{Dynamics-Aware Rollout:} We allocate stochastic exploration to dynamic regions and account for the policy-dependent mask in both sampling and Gaussian likelihoods.
    \item \textbf{Empirical Evaluation:} Training experiments demonstrate benchmark improvements across embodied world models, while complementary offline analyses assess reward reasonableness and resistance to erroneous rewards on hacking/control clips.
\end{itemize}

\par\endgroup

\section{Related Work}
\textbf{Embodied Video World Models.}
Recent progress has shifted video generation from passive visual synthesis toward \textbf{active} embodied world modeling, where generative and predictive models serve as interactive simulators of physical environments. Early systems such as DeepMind's Genie \cite{bruce2024geniegenerativeinteractiveenvironments} demonstrated the potential for learning controllable virtual worlds from large-scale unlabeled video. Related real-world simulators include GAIA-1 ~\cite{hu2023gaia1generativeworldmodel} and UniSim ~\cite{yang2023unisim}, focusing on autonomous driving and robotic manipulation. NVIDIA Cosmos~\cite{nvidia2026worldsimulationvideofoundation} develops video foundation models for physical AI, while V-JEPA~\cite{bardes2024vjepa} learns predictive visual representations and V-JEPA~2~\cite{assran2025vjepa2selfsupervisedvideo} extends these toward prediction and robotic planning. Complementary rectified-flow work~\cite{esser2024scaling} establishes scalable high-resolution image synthesis; it does not by itself demonstrate embodied physical compliance. By unifying action-conditioned prediction with scalable generative stability, these models are evolving into foundation world models that bridge the gap between high-fidelity simulation and downstream policy learning.
Building on this connection, world-action models further integrate visual dynamics prediction with robot action generation. WorldVLA unifies autoregressive image and action modeling~\cite{cen2025worldvla}, while DreamZero jointly predicts future video and actions using a pretrained video diffusion backbone~\cite{ye2026dreamzero}. LingBot-VA learns frame prediction and policy execution within an autoregressive diffusion framework~\cite{li2026lingbotva}. Cosmos Policy adapts a video model to generate actions, future states, and values for control and planning~\cite{kim2026cosmospolicy}. These approaches motivate our focus on evaluating the task evidence in predicted videos, so that reward-guided improvements target meaningful behavior.

\begingroup\color{black}\hypersetup{citecolor=black}
\textbf{Reward Models for Visual Generation.}
Reward-model design spans supervised quality prediction, preference learning, reasoning-based assessment, and model-derived physical signals. VideoScore~\cite{he2024videoscore} trains a multimodal evaluator on human annotations covering multiple aspects of video quality. VideoReward, released through VideoAlign~\cite{liu2025improvingvideogenerationhuman}, instead learns from multidimensional pairwise preferences over visual quality, motion quality, and text alignment. These designs provide learned scoring signals for evaluating or optimizing generated videos.

Reasoning-based models make the assessment process more explicit. VideoScore2~\cite{he2025videoscore2} produces rationales alongside scores for visual quality, text alignment, and physical/common-sense consistency, using supervised fine-tuning followed by GRPO. UnifiedReward-Flex~\cite{unifiedreward-flex} constructs context-dependent evaluation criteria under predefined and self-generated dimensions, combining distilled reasoning supervision with preference optimization. A complementary approach, WMReward~\cite{yuan2026wmreward}, uses a latent world model's physical prior to guide inference-time video generation. Thus, existing methods offer different forms of feedback, including scores, explanations, adaptive criteria, and physical evidence.

Our framework organizes such feedback within an extensible verification process. A reward model's score or rationale can be exposed through a tool adapter and checked against the instruction and video before influencing the final judgment. Our implementation uses WMReward together with motion tracking; the other reward models discussed here are potential extensions, not additional evaluated components. Their integration could broaden evaluation coverage, while explicit task requirements, evidence checks, and final gates remain responsible for determining whether positive credit is justified.
\endgroup

\textbf{Reinforcement Learning for World Model Post-Training.}
\label{sec:RL_worldmodel}
Recent post-training methods for generative world models increasingly adopt GRPO ~\cite{shao2024deepseekmathpushinglimitsmathematical} as a scalable alternative to Actor-Critic RL, enabling policy optimization without auxiliary value networks. Notably, Flow-GRPO ~\cite{liu2025flowgrpotrainingflowmatching} pioneered online RL in flow matching by converting deterministic ODE rollouts into stochastic SDE trajectories, establishing a critical balance between exploration and controllability. To address the noise artifacts induced by SDE-based stochasticity, Flow-CPS ~\cite{wang2025coefficientspreservingsamplingreinforcementlearning} reformulates the sampling process to maintain the coefficient-preserving property, thereby enabling more accurate reward modeling and stable convergence. Subsequent variants have largely refined this stability by enforcing rigorous distributional and sampling constraints. DanceGRPO~\cite{xue2025dancegrpounleashinggrpovisual} extends GRPO across visual generation tasks. UniGRPO~\cite{liu2026unigrpounifiedpolicyoptimization} jointly optimizes reasoning and image generation, replacing latent KL regularization with a velocity-field MSE penalty. MixGRPO~\cite{li2026mixgrpounlockingflowbasedgrpo} confines stochastic sampling and optimization to a sliding denoising window to improve efficiency, whereas Pref-GRPO~\cite{wang2026prefgrpopairwisepreferencerewardbased} uses pairwise preference rewards to mitigate pointwise reward hacking. Furthermore, frameworks such as SAGE-GRPO ~\cite{zheng2026sagegrpo} incorporate trust-region restrictions to stabilize policy updates, while systems built on Cosmos-Predict ~\cite{nvidia2026worldsimulationvideofoundation} often leverage temporal consistency priors to ensure physical plausibility. Beyond reverse-process policy gradients, DiffusionNFT~\cite{zheng2026diffusionnft} performs likelihood-free reinforcement through forward-process flow matching, using reward-weighted positive and negative objectives while allowing arbitrary sampling solvers. UniWorld-V2~\cite{li2025uniworldv2} extends this approach to image editing with multimodal-model feedback and group filtering to reduce scoring noise. These advances improve optimization efficiency and stability, but do not by themselves resolve reward hacking or ensure sampling diversity in embodied world models: policy updates still depend on the supplied reward, and solver flexibility does not guarantee diverse task-relevant interaction trajectories. This leaves reliable task verification and explicit rollout diversification as complementary needs, motivating our combination of Reward as an Agent and DynDiff-GRPO.

Recent on-policy distillation (OPD) methods address multi-reward alignment by consolidating specialized teachers. DiffusionOPD~\cite{li2026diffusionopd} trains task-specific teachers separately and distills them along student-generated trajectories; Flow-OPD~\cite{fang2026flowopd} further incorporates task routing and manifold anchoring. Although these designs mitigate cross-task interference, training and maintaining multiple teachers followed by a separate distillation stage complicates the training pipeline. Their objectives primarily integrate teacher capabilities rather than explicitly model correlations and dependencies among reward dimensions for the same generated video. Our agent instead jointly assesses task, physical, and visual evidence, with reward gates preventing favorable auxiliary scores from compensating for task failure.

\section{Reward as an Agent}

\subsection{Reward Hacking in Embodied World Model RL}
\label{sec:reward hacking}

We formulate denoising as a Markov Decision Process (MDP): the state contains the current latent, denoising time, and conditioning inputs; an action is the next latent sampled by the world-model policy. The completed video receives the task reward, while intermediate denoising rewards are zero:
\begin{equation}
\text{MDP} = (\mathcal{S}, \mathcal{A}, P, R), \qquad
J(\pi)=\mathbb{E}_{\pi}\Big[\sum_{t=0}^{T}R(s_t,a_t)\Big]
\end{equation}
Under this framework, policy improvement is fundamentally shaped by reward quality, since optimization follows the provided reward signal rather than the true objective itself.

\textbf{Reward Hacking Definition.}
This creates a critical vulnerability when reward functions are imperfect. Reward hacking occurs when the policy exploits flaws in proxy rewards rather than genuinely accomplishing the intended task~\cite{skalse2022rewardhacking}:
\begin{equation}
J(\pi_{\text{hack}}) > J(\pi_{\text{intended}}), 
\quad \text{while} \quad 
\text{TaskEval}(\pi_{\text{hack}}) \ll \text{TaskEval}(\pi_{\text{intended}})
\end{equation}
In other words, optimization succeeds mathematically but fails behaviorally. In embodied diffusion systems, this often manifests through exploiting visual shortcuts, shallow semantic alignment, or model artifacts while bypassing authentic physical interaction.

\textbf{Empirical Observation of Reward Hacking.}
This problem is particularly severe in embodied world models because embodied tasks require not only visual realism, but also semantic correctness, physical plausibility, and genuine task completion. However, most existing reward systems are inherited from general video generation and lack embodied-specific evaluation capability. To systematically examine this mismatch, we optimize embodied world models in embodied scenarios using DanceGRPO~\cite{xue2025dancegrpounleashinggrpovisual} with two representative general-purpose reward metrics, VideoAlign~\cite{liu2025improvingvideogenerationhuman} and UnifiedReward-Flex~\cite{unifiedreward-flex}. Despite continuous optimization, we observe that reward values largely fluctuate within a narrow range while downstream generation quality shows little meaningful improvement. To understand this discrepancy, we conduct detailed analysis of model rollouts alongside their corresponding reward signals, and identify reward-hacking cases in which high proxy scores do not reflect task success. Detailed scoring prompts and evaluation criteria are provided in Appendix~\ref{sec:appendix_case}.

\begin{figure}[t]
  \centering
  \includegraphics[width=1\textwidth]{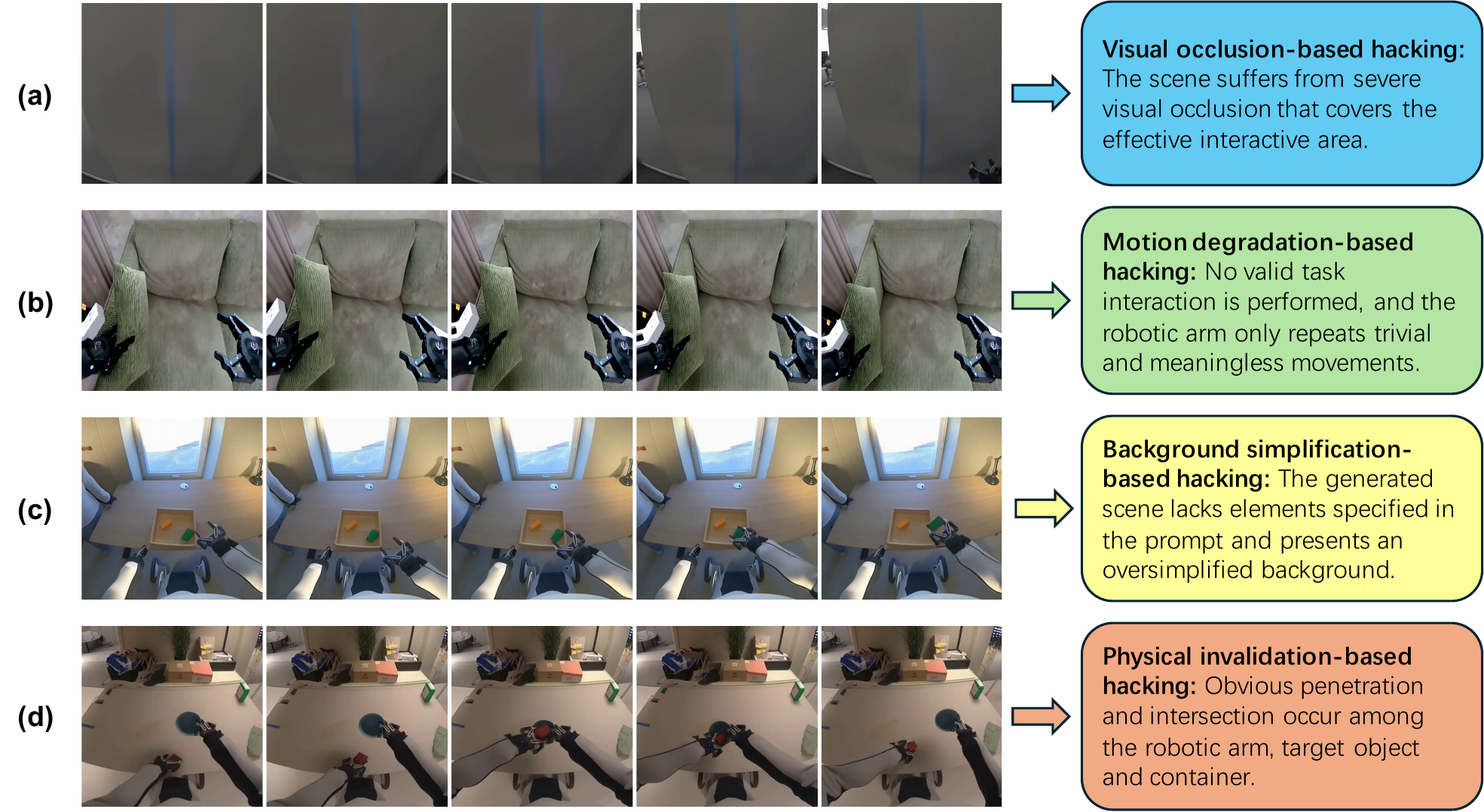}
  \caption{Typical reward hacking failures under existing reward metrics, where generated videos receive deceptively high scores despite violating true task objectives. 
  (a) Visual occlusion: $0.79/1.0$ on VideoAlign VQ; 
  (b) Motion degradation: $2.76/5.0$ on UnifiedReward-Flex Alignment Score; 
  (c) Background simplification: $0.80/1.0$ on VideoAlign VQ; 
  (d) Physical invalidation: $3.0/5.0$ on UnifiedReward-Flex Physics Score. 
  These cases highlight systematic blind spots in current reward designs.}
  \label{fig:reward_hacking}
\end{figure}

Our analysis reveals four recurring reward hacking patterns (Figure~\ref{fig:reward_hacking}): (1) \textbf{Visual occlusion}, where blur, shadows, or occlusion conceal physical defects; (2) \textbf{Motion degradation}, where the model minimizes risk through static or trivial movement; (3) \textbf{Background simplification}, where environmental complexity is reduced to avoid generation failures; and (4) \textbf{Physical invalidation}, where outputs superficially satisfy prompts while violating physical laws.

These observations reveal that current reward paradigms fail not because RL optimization itself is ineffective, but because the reward functions guiding optimization are structurally incomplete. Existing systems suffer from four major limitations: (1) perceptual metrics built on traditional computer vision detection models often focus on object recognition, trajectory matching, or pixel-level similarity, and do not directly assess embodied physical interactions; (2) VLM-based evaluators fine-tuned for specific domains may perform well on narrow tasks, but generally exhibit poor domain generalization when transferred beyond their specialized settings; (3) incomplete evaluation dimensions that overlook critical factors such as physical compliance or authentic task completion; and (4) reward aggregation that allows visual or physical scores to compensate for established task failure. As a result, policies are incentivized to maximize proxy scores through exploitable shortcuts rather than genuine embodied competence.

These failures suggest that reward hacking is fundamentally a reward design problem rather than merely an optimization problem: if reward functions remain unreliable, stronger RL algorithms may simply exploit them more effectively. The core challenge is therefore to redesign reward as a robust evaluative system that better aligns proxy objectives with true embodied goals.

\begingroup\color{black}\colorlet{themecolor}{black}\hypersetup{linkcolor=black,citecolor=black,urlcolor=black}
\subsection{Reward Design}
\label{sec:reward_design}
\textbf{Design through Iterative Evaluation.}
Our agent aims to verify task progress, reject unsupported credit, and preserve useful partial execution. Inspired by recursive self-improvement~\cite{zhang2025darwingodel} and automated agent design~\cite{hu2024adas}, we iteratively evaluated rewards and explanations on 100 ordinary and 80 hacking/control videos, then revised prompts, tools, and decision rules. This process produced 16 development versions, distinct from the 14 final-design ablations in Sec.~\ref{sec:reward_analysis}. These development-exposed samples do not establish held-out generalization.

Reward as an Agent combines a multimodal model with CoTracker3~\cite{karaev2025cotracker3} and WMReward~\cite{yuan2026wmreward} (Figure~\ref{fig:reward_agent}). The model gathers evidence and revises a structured assessment; fixed rules convert the final assessment into a reward.

\begin{figure}[t]
  \centering
  \includegraphics[width=\textwidth]{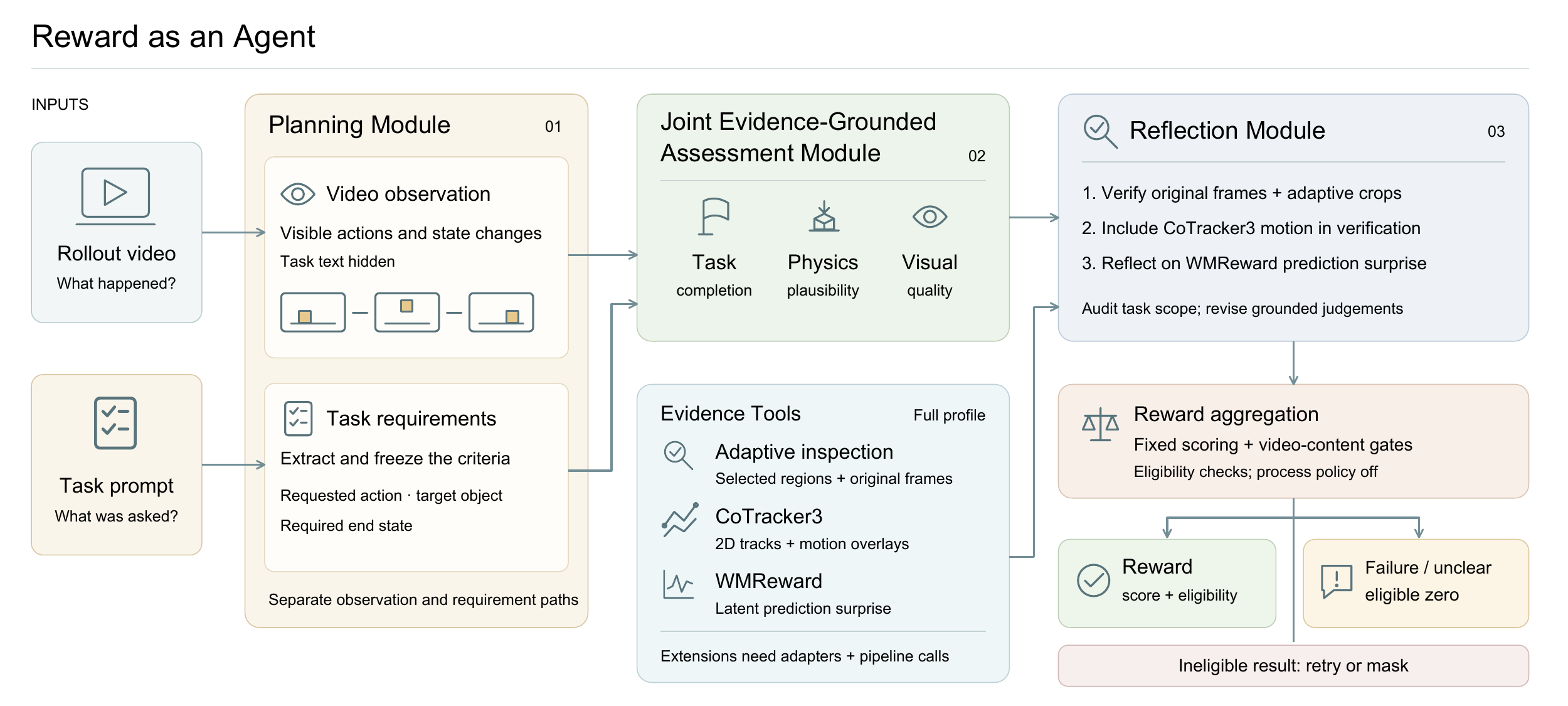}
  \captionsetup{font={color=black},labelfont={color=black}}
  \caption{Reward Agent workflow. Separate observations and task requirements support a joint assessment, followed by visual verification, tool-assisted reflection, and requirement repair. Tools provide evidence; final rules determine the reward and training eligibility.}
  \label{fig:reward_agent}
\end{figure}

\textbf{Separate Observation and Task Assessment.}
The agent first observes 32 uniformly sampled frames without the task instruction. It then extracts and freezes the instruction's required actions, objects, ordering, and outcomes as a \emph{task contract}. Separating these steps reduces the risk of mistaking desired behavior for observed behavior. Given the contract and video evidence, the model assesses \emph{task completion}, \emph{physical plausibility}, and \emph{visual quality}, recording requirement-level judgments and supporting frame references. Useful progress can receive partial credit; mere approach or ineffective contact does not establish success.

\textbf{Verify and Reflect with Tools.}
The agent revisits uncertain judgments using original frames, selected crops, and CoTracker3 motion overlays. It reviews all frames of clips with at most 81 frames; longer clips add up to 16 nearby frames to the initial sample. Invalid crop plans fall back to full-frame views. CoTracker3 supplies point trajectories and visibility, while WMReward uses V-JEPA~2~\cite{assran2025vjepa2selfsupervisedvideo} to provide latent prediction-surprise feedback for a subsequent visual reflection. Both tools supply supporting evidence: tracks do not certify contact or success, and prediction surprise is neither a physical-validity guarantee nor a direct reward term.

\textbf{Check Requirements and Resolve Failures.}
A scope audit checks missing requirements, invented conditions, and contradictions, allowing up to three image-backed repair rounds. Text-only suggestions cannot override visual judgments. When needed, a final visual check determines whether task failure is established despite remaining ambiguity. Required-tool and protocol failures make the evaluation ineligible, rather than triggering model-only fallback. An initial quality check may assign zero to unusable or spatially uniform video content before tool inference.

\textbf{Extend the Evidence Tools.}
Additional reward models, task-specific evaluators, or rule-based checks can be integrated through custom adapters and explicit verification or reflection calls. Their feedback must be checked against the task contract and visual evidence before revising the assessment. Our implementation evaluates CoTracker3 and WMReward; it does not provide automatic tool discovery or a generic plug-in registry.

\textbf{Reward Aggregation and Eligibility.}
Before target-mismatch adjustment, we map the task, physical, and visual judgments to fixed scores: $T\in\{1,0.8,0.45,0\}$ for complete, mostly complete, partial, and failed execution; $P\in\{1,0.75,0.25\}$ for plausible motion, minor defects, and major defects; and $V\in\{1,0.8,0.3\}$ for clear, mildly degraded, and severely degraded but assessable imagery. For assessable components, the diagnostic reward is
\begin{equation}
 R_{\mathrm{diag}}(v,c)=
 \bigl(0.70T+0.20P+0.10V\bigr)\bigl(0.5+0.5P\bigr).
 \label{eq:diagnostic_reward}
\end{equation}
An explicitly mismatched target caps $T$ at $0.2$. These mappings and weights are fixed design choices.

Let $E$ indicate that the evaluation passes the required execution and consistency checks, $F$ denote evidence-supported task failure, and $U$ denote video-content uncertainty that prevents justified positive credit. The final reward is
\begin{equation}
R(v,c)=\begin{cases}
\bot, & E(v,c)=0,\\
0, & E(v,c)=1\ \text{and}\ (F(v,c)=1\ \text{or}\ U(v,c)=1),\\
R_{\mathrm{diag}}(v,c), & \text{otherwise}.
\end{cases}
\label{eq:current_reward_gate}
\end{equation}
Supported partial progress retains graded credit, while failure or unverifiable essential content receives zero. Uncertainty-based zero does not imply confirmed failure. Here $\bot$ denotes an invalid evaluation that must be retried or excluded from training; only explicitly eligible results support policy updates. The full Agent uses API-based inference and local tools; runtime settings and an alternative local deployment appear in Appendix~\ref{sec:reward_service}.
\par\endgroup

\begingroup\color{black}\colorlet{themecolor}{black}\hypersetup{linkcolor=black,citecolor=black,urlcolor=black}
\section{Dynamic-Aware Rollout Diversification}
\label{sec:dar}
\subsection{Dynamic-Space Under-Exploration in Conservative RL Rollout}
Existing RL methods for diffusion world models are often overly conservative, constraining rollouts near the training distribution to avoid reward hacking (Sec.~\ref{sec:RL_worldmodel}). While this ensures stability, it severely limits the exploration of diverse dynamics. Crucially, in embodied tasks, policy improvement requires exploring multiple plausible dynamic evolutions within the same scene, even as static elements like environment geometry remain consistent. Standard conservative GRPO fails to distinguish between these dimensions; by uniformly suppressing rollout stochasticity, it collapses diffusion sampling toward deterministic, ODE-like trajectories. This leads to a narrow set of reward-favored behaviors, increasing reward overfitting and stifling the discovery of rich dynamic variety.

Therefore, effective embodied rollout diversification should not simply increase global randomness, but instead preserve scene fidelity while expanding dynamic-space exploration. Motivated by this, we propose \textbf{DynDiff-GRPO}, a dynamic-aware stochastic rollout strategy that selectively reallocates stochasticity toward dynamically salient regions while maintaining stable scene consistency.

\subsection{DynDiff-GRPO: Dynamic-Aware Stochastic Rollout}
\textbf{Dynamic-Aware Stochastic Rollout Diversification.}
Given latent state $x_t$ and velocity prediction $v_\theta(x_t,t,c)$, we first estimate the predicted clean sample and noise component:
\begin{equation}
\hat{x}_0 = x_t - \sigma_t v_\theta, \qquad
\hat{\epsilon}_t = x_t + (1-\sigma_t)v_\theta.
\label{eq:x0_epsilon}
\end{equation}

Building on the coefficient-preserving sampling analysis of CPS~\cite{wang2025coefficientspreservingsamplingreinforcementlearning}, which theoretically shows that reducing excessive stochastic perturbation during reverse diffusion can yield a more stable and effective SDE rollout process than conventional flow-SDE sampling, we further generalize this insight into a hybrid ODE/SDE transition framework. Specifically, rather than fixing rollout dynamics, we continuously interpolate between deterministic exploitation and stochastic exploration:

\begin{equation}
x_{t^-} =
\hat{x}_0(1-\sigma_{t^-}) +
\hat{\epsilon}_t
\sqrt{
\sigma_{t^-}^2-
(\sigma_{t^-}^{\text{noise}})^2
}
+
\sigma_{t^-}^{\text{noise}}
\odot z,
\quad
z\sim\mathcal{N}(0,I).
\label{eq:x_t_minus}
\end{equation}

where $\sigma_{t^-}^{\text{noise}} \in [0,\sigma_{t^-}]$ controls exploration strength, recovering deterministic ODE rollout when $\sigma_{t^-}^{\text{noise}}=0$ and standard stochastic rollout when $\sigma_{t^-}^{\text{noise}}=\sigma_{t^-}$. Instead of uniformly distributing stochasticity across spatial locations, DynDiff-GRPO reallocates exploration according to dynamic saliency:

For frame index $k$, we construct the noise scale and dynamic priors as
\begin{align}
\sigma_{t^-}^{\text{noise}} &= \sigma_{t^-}\odot M_t,
& R^{(k)} &= \Delta\hat{x}_0^{(k)}-\mathrm{Mean}_{h,w}\!\left(\Delta\hat{x}_0^{(k)}\right),
\label{eq:sigma_noise}\\
D &= \mathrm{Normalize}(\|R\|_2),
& B &= \mathbb{I}[D>\mathrm{Quantile}(D,\tau)].
\label{eq:dynamic_intensity}
\end{align}
Here, $\Delta\hat{x}_0^{(k)}$ is the difference between adjacent predicted frames. Subtracting its spatial mean isolates localized dynamics; normalization and percentile thresholding yield the continuous intensity $D$ and sparse prior $B$.

Finally, continuous dynamic intensity and sparse dynamic-region priors are combined to produce the rollout mask:

\begin{equation}
M_t
=
r_{\text{base}}
+
(1-r_{\text{base}})
\left(
0.5D + 0.5B
\right),
\label{eq:dynamic_mask}
\end{equation}

where $r_{\text{base}}$ defines the minimum exploration ratio, ensuring that low-dynamic regions preserve a non-zero level of stochasticity while highly dynamic regions receive stronger exploration.

Substituting the adaptive \textcolor{black}{variance} into Eq.~\ref{eq:x_t_minus}, the rollout transition distribution becomes:

\par\noindent
\begin{minipage}{.38\linewidth}
\begin{equation}
x_{t^-}
\sim
\mathcal{N}
\left(
\mu,
(\sigma_{t^-}\odot M_t)^2
\right),
\label{eq:rollout_dist}
\end{equation}
\end{minipage}\hfill
\begin{minipage}{.60\linewidth}
\begin{equation}
\mu=\hat{x}_0(1-\sigma_{t^-})+\hat{\epsilon}_t\sqrt{\sigma_{t^-}^2-(\sigma_{t^-}^{\text{noise}})^2}.
\label{eq:rollout_mean}
\end{equation}
\end{minipage}
\par

The following expression gives the element-wise log density; summing over latent coordinates yields the transition log-likelihood used during policy optimization:

\begin{equation}
\log p(x_{t^-})
=
-\frac{
(x_{t^-}-\mu)^2
}{
2(\sigma_{t^-}\odot M_t)^2
}
-
\log(\sigma_{t^-}\odot M_t)
-
\log\sqrt{2\pi}.
\label{eq:rollout_logprob}
\end{equation}

See Sec.~\ref{sec:rollout_analysis} for quantitative analysis.

\textbf{Group-Relative Policy Optimization for Diffusion World Models.}
Using the dynamic-aware stochastic sampler, for each conditioning prompt $c$, DynDiff-GRPO samples a rollout group $\{\tau_i\}_{i=1}^{G}\sim \pi_{\theta_{\mathrm{old}}}(\cdot|c)$ under independently diversified diffusion paths, producing rewards $r_i=R(\tau_i,c)$.
Let $\mathcal{I}=\{i:E(\tau_i,c)=1\}$ be the eligible rollout indices and $n=|\mathcal{I}|$. Content-based uncertainty can yield an eligible zero; evaluator faults are retried or excluded. Groups with fewer than two eligible rollouts are skipped. For $i\in\mathcal{I}$, we use clipped group-relative advantages:
\begin{equation}
\begin{gathered}
\bar r=\frac{1}{n}\sum_{j\in\mathcal{I}}r_j,\qquad
s_r=\sqrt{\frac{1}{n}\sum_{j\in\mathcal{I}}(r_j-\bar r)^2},\\
A_i=\mathrm{clip}\!\left(\frac{r_i-\bar r}{s_r+\epsilon},-A_{\max},A_{\max}\right).
\end{gathered}
\label{eq:advantage}
\end{equation}
Here $\epsilon>0$ stabilizes normalization; equal-reward groups yield zero advantages. Dynamic awareness enters the rollout mask and policy likelihood, while reward normalization uses the eligible group's standard deviation.

For each denoising step, we model the reverse transition as a Gaussian policy.
\begingroup\color{black}
\textbf{Policy-Dependent Variance and Likelihood.}
CPS~\cite{wang2025coefficientspreservingsamplingreinforcementlearning} simplifies its likelihood ratio using a noise scale shared by the current and old policies. In DynDiff-GRPO, the mask is constructed from the model's predicted clean sample, coupling dynamic-aware exploration to the transition distribution used for policy optimization. Evaluating each policy's mask from its own prediction yields the feature-wise standard deviation $s_{\theta,t}=\sigma_{t^-}\odot M_{\theta,t}(x_t,c)$. The Gaussian log-likelihood is
\begin{equation}
\log \pi_\theta(x_{t^-}|x_t,c)=
-\sum_j\left[
\frac{(x_{t^-,j}-\mu_{\theta,j})^2}{2s_{\theta,t,j}^2}
+\log s_{\theta,t,j}
+\frac{1}{2}\log(2\pi)
\right],
\label{eq:logpi}
\end{equation}
where $j$ indexes latent coordinates. This density is defined on stochastic steps $\mathcal{T}_{+}=\{t:\sigma_{t^-}>0\}$, with $0<r_{\mathrm{base}}\leq1$ ensuring positive coordinate-wise scales. A deterministic terminal transition with $\sigma_{t^-}=0$ is excluded from likelihood ratios and the policy loss.

For the same sampled transition, the current and old policies can have different dynamic masks and hence different standard-deviation fields. The importance ratio must therefore retain both \textbf{variance-weighted residuals and log-scale normalization}. The Gaussian constant cancels at fixed dimensionality, whereas the log-scale contribution $\sum_j\log(s_{\theta_{\mathrm{old}},t,j}/s_{\theta,t,j})$ generally does not. If both policies share a fixed mask, the log-scale terms cancel, but the residuals remain weighted by the spatial variance.

This couples dynamic-aware exploration with GRPO updates that account for changes in both the transition mean and scale. We compute the importance ratio using each policy's corresponding Gaussian likelihood:
\par\endgroup
\begin{equation}
\rho_{i,t}=
\exp\!\left(
\log \pi_\theta(x_{t^-}^{(i)}|x_t^{(i)},c)
-
\log \pi_{\theta_{\mathrm{old}}}(x_{t^-}^{(i)}|x_t^{(i)},c)
\right).
\label{eq:ratio}
\end{equation}

The policy objective follows GRPO~\cite{shao2024deepseekmathpushinglimitsmathematical}, using the clipped probability-ratio surrogate introduced in PPO~\cite{schulman2017ppo}, with the expectation restricted to $i\in\mathcal{I}$ and $t\in\mathcal{T}_{+}$:
\begin{equation}
\mathcal{L}_{\mathrm{GRPO}}=
-\mathbb{E}_{i,t}\!\left[
\min\!\left(
\rho_{i,t}A_i,\,
\mathrm{clip}(\rho_{i,t},1-\delta,1+\delta)A_i
\right)
\right].
\label{eq:grpo}
\end{equation}

For regularization, following CPS and SAGE-GRPO, we constrain policy updates toward the target distribution with a mean-space KL-style regularizer and combine it with the GRPO objective:
\par\noindent
\begin{minipage}{.48\linewidth}
\begin{equation}
\mathcal{L}_{\mu\text{-KL}}
=
\left\|
\frac{
\mu_\theta - \mu_{\mathrm{target}}
}{
\sigma_{\mathrm{target}} + \epsilon
}
\right\|_2^2.
\label{eq:mu_kl}
\end{equation}
\end{minipage}\hfill
\begin{minipage}{.48\linewidth}
\begin{equation}
\mathcal{L}
=
\mathcal{L}_{\mathrm{GRPO}}
+
\beta \mathcal{L}_{\mu\text{-KL}}.
\label{eq:total_loss}
\end{equation}
\end{minipage}
\par

Through dynamic-aware rollout diversification, dynamic-aware vectorized Gaussian likelihood estimation,
clipped group-relative optimization, and mean-space regularization,
DynDiff-GRPO transforms expanded dynamic-space exploration into stable and reward-aligned diffusion policy improvement. See Appendix~\ref{sec:DynDiffgrpo_code} for rollout implementation details.

\par\endgroup

\section{Experimental Results}
\label{sec:experiment}
\begingroup\color{black}\colorlet{themecolor}{black}\hypersetup{linkcolor=black,citecolor=black,urlcolor=black}
\captionsetup{font={color=black},labelfont={color=black}}
\subsection{Main Experiment}
\textbf{Evaluation settings.} We evaluate on PAI-Bench's robotics subset (174 tasks, 913 VQA questions)~\cite{zhou2025paibenchcomprehensivebenchmarkphysical}. On a 0--100 scale, Domain is question-level accuracy, Quality averages the eight visual metrics in Table~\ref{tab:paibench_robot}, and Overall averages both. All models are evaluated using Qwen3.8-Flash-Next with eight uniformly sampled frames, fixed binary prompts, and greedy decoding (temperature 0, top-p 1, thinking disabled); repeated scoring of identical videos under the same configuration reproduces the results.

\textbf{Experimental setup.} We train Cosmos-Predict2.5-2B~\cite{nvidia2026worldsimulationvideofoundation} and Kairos-3.0-Robot (Mar. Version)~\cite{kairosteam2026kairosnativeworldmodel}, with supplementary Kairos3.1 (4B) experiments using the Kairos3.0 configuration. All runs use Doubao 2.1 Pro as the Reward Agent backbone (Sec.~\ref{sec:reward_analysis}). Training settings are listed in Appendix~\ref{sec:training_configurations}, with configuration analyses in Sec.~\ref{sec:analysis}.

\textbf{Training data.} Public training data are sampled to match PAI-Bench's robot-embodiment proportions, with sequences, initial frames, and prompts disjoint from the test set. Only the initial frame and an augmented task prompt condition each rollout; subsequent source frames are not training targets.

\begin{table}[t]
\centering
\footnotesize
\setlength{\tabcolsep}{3pt}
\renewcommand{\arraystretch}{1.08}
\caption{
Performance on the robotics subset of PAI-Bench.
Numbers in parentheses indicate absolute improvements over the corresponding base model. Bold values improve over the respective baseline. All models and their DynDiff-GRPO counterparts use the same Qwen Flash evaluator and evaluation settings.
}
\label{tab:paibench_robot}

\resizebox{\columnwidth}{!}{
\begin{tabular}{llcccccccc|cc}
\toprule
Model & Param &
i2v-bg & i2v-s & aes & img &
bg-con & mot & sub-con & o-con &
Domain $\uparrow$ & Overall $\uparrow$ \\
\midrule

Cosmos-Predict2.5 & 2B
& 93.47 & 91.13 & 47.74 & 70.45
& 90.86 & 98.80 & 90.06 & 19.30
& 83.61 & 79.42 \\

DynDiff-GRPO (Ours) & 2B
& \textbf{96.48} & \textbf{96.97} & \textbf{49.14} & \textbf{70.60}
& 90.47 & \textbf{99.32} & \textbf{91.99} & \textbf{19.68}
& \textbf{84.41} (+0.80)
& \textbf{80.62} (+1.20) \\

\midrule

Kairos3.0-Robot & 4B
& 97.75 & 95.00 & 46.72 & 70.78
& 92.31 & 99.26 & 93.32 & 19.56
& 83.87 & 80.35 \\

DynDiff-GRPO (Ours) & 4B
& 97.17 & \textbf{96.05} & \textbf{46.94} & \textbf{71.32}
& \textbf{92.95} & 99.25 & 91.11 & \textbf{19.68}
& \textbf{86.95} (+3.08)
& \textbf{81.88} (+1.53) \\

\midrule

Kairos3.1 & 4B
& 99.22 & 95.06 & 45.40 & 67.54
& 94.90 & 99.47 & 95.71 & 19.46
& 85.43 & 81.26 \\

DynDiff-GRPO (Ours) & 4B
& \textbf{99.44} & \textbf{96.12} & \textbf{46.83} & \textbf{70.16}
& 94.87 & 99.22 & 95.63 & \textbf{19.67}
& \textbf{88.31} (+2.88) & \textbf{83.03} (+1.77) \\

\bottomrule
\end{tabular}
}
\end{table}

\textbf{Main Results.}
Table~\ref{tab:paibench_robot} shows Domain gains of 0.80, 3.08, and 2.88 points on Cosmos, Kairos3.0, and Kairos3.1, with Overall gains of 1.20, 1.53, and 1.77. Kairos3.1 reaches 88.31 Domain and 83.03 Overall; its mean Quality increases by 0.65 to 77.74, with five of eight metrics improving. Gains are computed relative to each model's corresponding baseline under the same evaluation settings.

\par\endgroup

\begingroup\color{black}\colorlet{themecolor}{black}\hypersetup{linkcolor=black,citecolor=black,urlcolor=black}
\subsection{Reward Analysis}
\label{sec:reward_analysis}
\textbf{Setup.}
We evaluate 14 configurations per backbone (Doubao 2.1 Pro and Qwen3.8-Flash-Next) on 180 videos: 100 ordinary and 80 hacking/control clips. Nine configurations cumulatively extend direct scoring; five remove individual modules from the full agent, preserving unaffected evidence and downstream settings. We also evaluate VideoAlign~\cite{liu2025improvingvideogenerationhuman} and UnifiedReward-Flex~\cite{unifiedreward-flex} as standalone reward baselines. Direct scoring receives only the video, instruction, and scoring criteria. Identical stage inputs share outputs within each evaluation run.

\textbf{Reward Reasonableness.}
Doubao 2.1 Pro~\cite{bytedance2026seed21} judges both backbones using the video, instruction, reward, and explanation, with method identities hidden, up to 64 frames, and temperature 0. It accepts scores consistent with task progress and execution quality without decisive reasoning errors, while tolerating minor calibration differences. Unverifiable essential content warrants zero; execution errors are retried. We report the accepted fraction of 180 outputs (prompt in Appendix~\ref{sec:reward_judge_exact_prompt}).

\begin{table}[H]
\centering\color{black}\small
\captionsetup{font={small,color=black},labelfont={color=black}}
\setlength{\tabcolsep}{3pt}
\caption{Reward reasonableness and hacking/control errors. R: reasonable judgments among 180 videos (Doubao 2.1 Pro judge); F+: positive rewards on 60 reference-negative clips; P0: zero rewards on 20 progress clips. Standalone rewards use their own models; Agent columns use Doubao 2.1 Pro and Qwen Flash. Modules are added cumulatively (middle) or removed individually (bottom). Bold: best per backbone, including standalone baselines.}
\label{tab:reward_system_eval_matrix}
\label{tab:reward_hacking_suppression}
\resizebox{\linewidth}{!}{
\begin{tabular}{lcccccc}
\toprule
Method & \multicolumn{2}{c}{R / 180 (\%) $\uparrow$} & \multicolumn{2}{c}{F+ / 60 $\downarrow$} & \multicolumn{2}{c}{P0 / 20 $\downarrow$} \\
\midrule
VideoAlign~\cite{liu2025improvingvideogenerationhuman} & \multicolumn{2}{c}{35 (19.4\%)} & \multicolumn{2}{c}{60} & \multicolumn{2}{c}{\textbf{0}} \\
UnifiedReward-Flex~\cite{unifiedreward-flex} & \multicolumn{2}{c}{42 (23.3\%)} & \multicolumn{2}{c}{60} & \multicolumn{2}{c}{\textbf{0}} \\
\midrule
Configuration & Doubao & Qwen & Doubao & Qwen & Doubao & Qwen \\
\midrule
Direct base-model scoring & 132 (73.3\%) & 115 (63.9\%) & 50 & 57 & 1 & \textbf{0} \\
+ Structured report and score mapping & 140 (77.8\%) & 140 (77.8\%) & 47 & 52 & 1 & 3 \\
+ Task decomposition & 156 (86.7\%) & 138 (76.7\%) & 53 & 52 & 1 & 4 \\
+ Task-blind observation & 170 (94.4\%) & 145 (80.6\%) & 42 & 55 & 2 & 5 \\
+ Adaptive visual verification & 169 (93.9\%) & 146 (81.1\%) & 35 & 42 & 1 & 2 \\
+ CoTracker3 motion evidence & 170 (94.4\%) & 148 (82.2\%) & 37 & 47 & 3 & 3 \\
+ WMReward feedback and reflection & 172 (95.6\%) & 146 (81.1\%) & 40 & 52 & 3 & 3 \\
+ Scope audit and repair (without failure gate) & 174 (96.7\%) & 152 (84.4\%) & 48 & 51 & 3 & 1 \\
\midrule
Full Reward Agent & \textbf{178 (98.9\%)} & \textbf{159 (88.3\%)} & \textbf{0} & 3 & \textbf{0} & 1 \\
Without task-blind observation & 169 (93.9\%) & 146 (81.1\%) & 6 & 6 & 2 & \textbf{0} \\
Without adaptive visual verification & 169 (93.9\%) & 138 (76.7\%) & 5 & 3 & 4 & 4 \\
Without WMReward feedback and reflection & 171 (95.0\%) & 142 (78.9\%) & 5 & 4 & 4 & 4 \\
Without scope audit and repair & 174 (96.7\%) & 141 (78.3\%) & 5 & \textbf{2} & 4 & 5 \\
Without CoTracker3 motion evidence & 170 (94.4\%) & 146 (81.1\%) & 6 & 3 & 4 & 3 \\
\bottomrule
\end{tabular}
}
\end{table}

\textbf{Results.}
The full agent achieves the highest reasonableness: 98.9\% for Doubao and 88.3\% for Qwen Flash, exceeding direct scoring by 25.6 and 24.4 percentage points. Individual module removals, including the failure gate, reduce reasonable judgments by 4--9 and 7--21, respectively, indicating complementary benefits. VideoAlign and UnifiedReward-Flex achieve 19.4\% and 23.3\% reasonableness; both assign positive rewards to all 60 reference-negative clips, despite producing no false zeros on the 20 progress clips.

\textbf{Hacking/control evaluation.}
The 80 clips comprise 57 failures, 3 unobservable cases, and 20 with partial or complete progress, according to automatic reference labels withheld from the judge. These labels are not human ground truth. Table~\ref{tab:reward_hacking_suppression} reports positive rewards on the first 60 clips (F+) and zero rewards on the latter 20 (P0).

The failure gate drives the largest reduction in erroneous positive rewards: with all other modules fixed, F+ falls from 48/60 to 0/60 for Doubao and from 51/60 to 3/60 for Qwen. P0 also improves from 3/20 to 0/20 for Doubao and remains 1/20 for Qwen, showing that this reduction does not sacrifice supported progress. Other modules help balance rejection and progress: removing scope audit lowers Qwen's F+ from 3 to 2 but raises P0 from 1 to 5 and lowers reasonableness to 78.3\%. The Agent supports both the closed-source Doubao 2.1 Pro and the open-source Qwen Flash backbone. We select Doubao 2.1 Pro for the final reward system and the training experiments in Table~\ref{tab:paibench_robot}, based on its higher reasonableness and zero errors on both hacking/control metrics in this evaluation.

\par\endgroup

\begingroup\color{black}\colorlet{themecolor}{black}\hypersetup{linkcolor=black,citecolor=black,urlcolor=black}
\captionsetup{font={color=black},labelfont={color=black}}
\subsection{Rollout Strategy Analysis}
\label{sec:rollout_analysis}
We compare DynDiff-GRPO with CPS~\cite{wang2025coefficientspreservingsamplingreinforcementlearning} through rollout diversity and PAI-Bench ablations.
\textbf{Rollout Diversity Analysis.}
For each video, we concatenate $L_2$-normalized DINOv3~\cite{Simeoni2025DINOv3} features from $T$ uniformly sampled frames, then normalize the concatenation to obtain $\hat{z}_v$. Diversity across $N$ rollouts per prompt is their mean pairwise cosine distance:
\begin{equation}
D_{\text{prompt}} = \frac{2}{N(N-1)} \sum_{1 \le i < j \le N} \left(1 - \hat{z}_i^\top \hat{z}_j\right).
\end{equation}
Figure~\ref{fig:dyndiff_vs_CPS}(a) shows that DynDiff-GRPO consistently maintains higher diversity than CPS during training, \textcolor{black}{indicating greater diversity in this feature space; this metric alone does not establish physical validity.}

\textbf{Robotics Evaluation on PAI-Bench.}
All variants use Kairos-3.0-Robot~\cite{kairosteam2026kairosnativeworldmodel}, identical training data, and $480 \times 832$ resolution. Reward Agent, training steps, rollout counts, CFG, and KL regularization are fixed; only the rollout strategy varies. 
Figure~\ref{fig:dyndiff_vs_CPS}(b) compares CPS, shifted-mask DynDiff-GRPO, and full DynDiff-GRPO. Relative to CPS, the full method improves Domain by 1.60 points (85.63 vs.\ 84.03) while maintaining similar Quality (76.74 vs.\ 76.58).
\begin{figure}[t]
  \centering
  \includegraphics[width=0.85\linewidth]{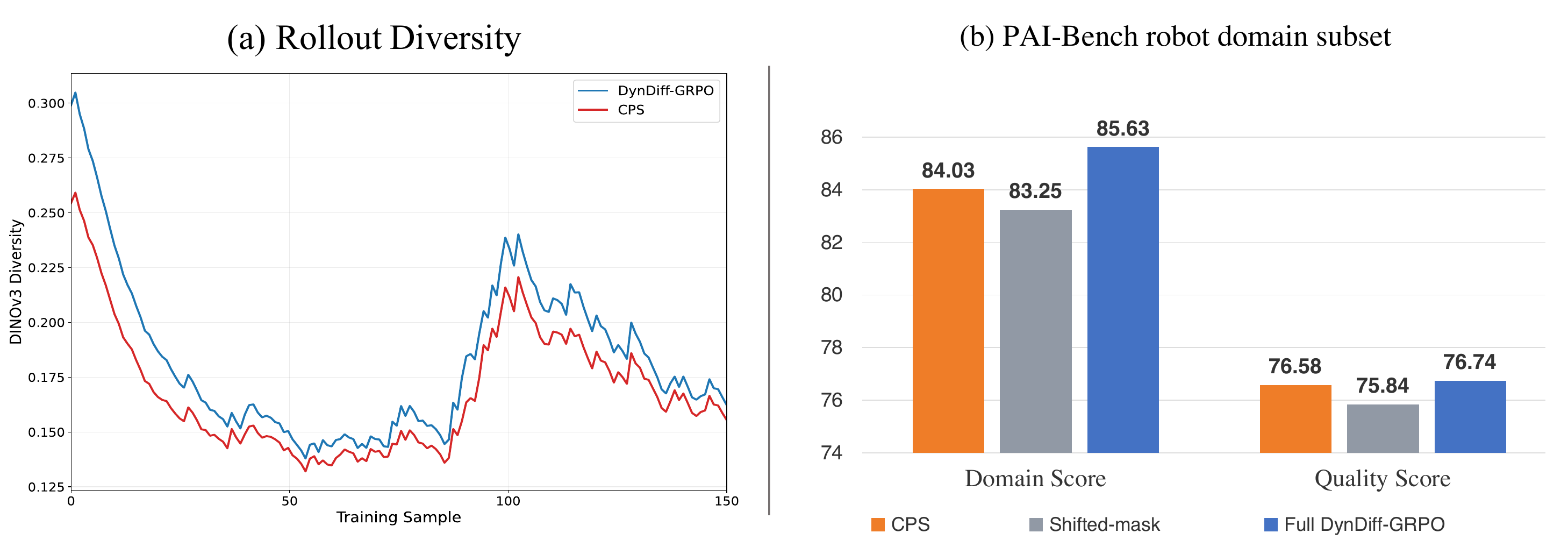} 
  
  \caption{{Ablation of rollout strategies.} (a) Rollout diversity via mean pairwise DINOv3 cosine distance. (b) Domain and Quality scores on the PAI-Bench robot domain subset for CPS, shifted-mask DynDiff-GRPO, and full DynDiff-GRPO. Quality averages eight visual metrics; higher is better.}
  
  \label{fig:dyndiff_vs_CPS}
\end{figure}

\begingroup\color{black}
\textbf{Spatial-Alignment Ablation.}
We circularly shift each dynamic mask by half the latent height and width, preserving its value distribution but displacing exploration. This shift is fixed across frames and denoising steps in both sampling and likelihood evaluation. In Fig.~\ref{fig:dyndiff_vs_CPS}(b), full DynDiff-GRPO exceeds this control by 2.38 Domain points (85.63 vs.\ 83.25) and 0.90 Quality points (76.74 vs.\ 75.84). The control also underperforms CPS, supporting the benefit of aligning exploration with dynamic regions.
\par\endgroup

\par\endgroup

\begingroup\color{black}\colorlet{themecolor}{black}\hypersetup{linkcolor=black,citecolor=black,urlcolor=black}
\captionsetup{font={color=black},labelfont={color=black}}
\subsection{Analysis of Key Training Configurations}
\label{sec:analysis}
In addition to algorithmic design, we further investigate several configuration choices that have a substantial impact on training stability and final performance, including the KL regularization strength, advantage clipping strategy, input resolution, and classifier-free guidance (CFG) settings.

\textbf{KL Regularization Formulations.}
\textcolor{black}{The full Gaussian log-likelihood in Eq.~\ref{eq:logpi} is used for the GRPO importance ratio. Separately, Eq.~\ref{eq:mu_kl} supplies a variance-normalized mean-distance regularizer. When the current and target masks yield different variances, this mean-only penalty is a surrogate rather than the full Gaussian KL, which also contains variance-ratio and log-scale terms. Using the surrogate for regularization does not remove those terms from the policy likelihood.} Building upon this mean-space surrogate, we further introduce a memory- and compute-efficient implementation for periodic regularization. Conventional RLHF and GRPO frameworks typically maintain a frozen reference model in memory to compute the KL penalty $\mathcal{D}_{KL}(\pi_\theta \| \pi_{\text{ref}})$. For large-scale Diffusion Transformers (DiTs)~\cite{peebles2022dit}, this severely bottlenecks GPU memory and requires an additional computational forward pass during every optimization step. 

To address this, we completely eliminate the explicit reference model by repurposing the data buffer's lifecycle as a periodic anchor. During the rollout phase, the active policy weights $\theta_{\text{old}}$ generate the trajectories. We cache the distribution parameters, specifically the mean $\mu_{\theta_{\text{old}}}$ and standard deviation $\sigma_{\theta_{\text{old}}}$, directly into the rollout buffer. During the subsequent optimization phase, these cached values serve as the reference targets for the $\mu$-KL objective over a predefined number of update steps ($N = \text{kl\_period\_steps}$). Specifically, we instantiate the targets in Eq.~\ref{eq:mu_kl} as $\mu_{\mathrm{target}} = \mu_{\theta_{\text{old}}}$ and $\sigma_{\mathrm{target}} = \sigma_{\theta_{\text{old}}}$, yielding our buffer-anchored periodic KL penalty:
\begin{equation}
\mathcal{L}_{\mu\text{-KL}}
=
\mathbb{E}_{s \sim \mathcal{D}} 
\left[
\left\|
\frac{
\mu_\theta(s) - \mu_{\theta_{\text{old}}}(s)
}{
\sigma_{\theta_{\text{old}}}(s) + \epsilon
}
\right\|_2^2
\right].
\label{eq:mu_kl_periodic}
\end{equation}
This buffer-based anchoring strategy balances plasticity and stability while providing two critical systemic advantages:
\begin{itemize}
    \item \textbf{Reduced Reference-Model Memory:} By discarding the explicit frozen reference model, we entirely eliminate the GPU memory cost of hosting a second DiT network. This substantial memory saving allows for larger rollout batch sizes, higher input resolutions, or longer frame generation contexts under the same hardware constraints.
    \item \textbf{Enhanced Training Throughput:} Utilizing cached parameters bypasses the redundant forward pass that would otherwise be required by a reference model during the gradient update step. The periodic updating of the anchor ($\theta_{\text{old}}$) naturally restricts policy drift within each buffer lifecycle, achieving the stabilizing effects of a periodically updated reference model but at no additional reference-model forward pass.
\end{itemize}

\textbf{Advantage Clipping.}
\textcolor{black}{Advantage clipping limits extreme group-relative updates. We use a threshold of 2.5 in the reported training configurations (Appendix~\ref{sec:training_configurations}). Our reward design uses fixed categorical mappings and evidence gating; it does not use voting or curriculum-based score aggregation. Isolating the effect of the reward distribution on training stability would require a matched reward-design ablation.}

\textbf{Classifier-Free Guidance (CFG).}
We further study the role of CFG~\cite{ho2022classifierfreediffusionguidance} during rollout generation in diffusion transformer (DiT) training. As shown in Table~\ref{tab:kairos_cfg_ablation}, enabling CFG improves most quality metrics, leading to slightly higher domain and overall scores while maintaining comparable perceptual quality. This is consistent with prior diffusion modeling results, where CFG strengthens conditional guidance by amplifying conditional signals during sampling.
However, these gains come with additional sampling overhead due to extra forward evaluations, which reduces rollout throughput in large-scale RL training. In contrast, \textcolor{black}{disabling CFG removes these extra forward evaluations} while still retaining competitive final performance under well-shaped reward signals and proper KL regularization.
Overall, our results suggest that CFG is an effective auxiliary enhancement rather than a strictly necessary component. In practice, we often disable CFG under high-resolution training settings (e.g., $640\times820$ and above) to prioritize rollout efficiency while maintaining strong quantitative and perceptual performance.

\begin{table}[t]
\vspace{0.5\baselineskip}
\caption{Ablation study on Kairos3.0-Robot on the PAIbench robot domain subset. We compare DynDiff-GRPO with and without CFG under identical base model, data, and $480\times832$ resolution. CFG improves domain and overall performance while maintaining strong perceptual quality. Best scores are shown in \textbf{bold}.}
\label{tab:kairos_cfg_ablation}
\vspace{0.5\baselineskip}
\centering
\small
\setlength{\tabcolsep}{4.2pt}
\renewcommand{\arraystretch}{1.15}
\begin{tabular}{l c c c c c c c c c c}
\toprule
\multirow{2}{*}{Method} 
& \multicolumn{8}{c}{Quality score} 
& \multirow{2}{*}{\begin{tabular}[c]{@{}c@{}}Domain\\ score\end{tabular}} 
& \multirow{2}{*}{\begin{tabular}[c]{@{}c@{}}Overall\\ score\end{tabular}} \\
\cmidrule(lr){2-9}
& i2v-bg & i2v-s & aes & img & bg-con & mot & sub-con & o-con & & \\
\midrule
w/o CFG
& 97.13 
& 94.26 
& \textbf{46.55}
& \textcolor{black}{\textbf{69.55}}
& 91.84 
& 99.27 
& 92.58 
& \textbf{19.54}
& 85.07 
& 80.71 \\
w/ CFG
& \textbf{97.89}
& \textbf{95.03} 
& 46.47 
& 68.98
& \textbf{92.93} 
& \textbf{99.36} 
& \textbf{93.74}
& 19.50 
& \textbf{85.63} 
& \textbf{81.18} \\
\bottomrule
\end{tabular}
\vspace{0.5\baselineskip}
\end{table}

\textbf{Resolution Selection.}
We further study the role of resolution selection in RL-based training. In our RL setup, rollout resolution can be selected without requiring a matching-resolution target video. We therefore evaluate both training-resolution changes and inference-time scaling.

\textcolor{black}{As shown in Table~\ref{tab:kairos_avg_quality}, we first evaluate a low-resolution setting (480$\times$832) with CFG enabled and observe stable performance across both quality and domain scores. We then increase the training resolution to 640$\times$820 and further test inference-time scaling. Without additional training, raising the inference resolution to 720$\times$960 improves domain alignment and overall performance, although average Quality Score decreases from 77.24 to 76.81. These results suggest that RL-trained models can better exploit higher-resolution inference distributions even when such data is scarce during training, highlighting the benefit of resolution-flexible optimization for improving domain alignment and overall performance.}

\begin{table}[t]
\vspace{0.5\baselineskip}
\caption{Ablation study on Kairos3.0-Robot on the PAIbench robot domain subset. We report average quality score, domain score, and overall score under different training/inference resolutions and CFG settings. Best results are shown in \textbf{bold}.}
\label{tab:kairos_avg_quality}
\vspace{0.5\baselineskip}
\centering
\small
\setlength{\tabcolsep}{4pt}
\renewcommand{\arraystretch}{1.15}

\begin{tabular}{l c c c c c c}
\toprule
Method
& Train
& Infer
& CFG
& Quality (Avg.)
& Domain
& Overall \\
\midrule

Kairos-LR
& 480$\times$832
& 480$\times$832
& Yes
& 76.74
& 85.63
& 81.18 \\

Kairos-HR
& 640$\times$820
& 640$\times$820
& No
& \textbf{77.24}
& 85.6
& 81.42 \\

Kairos-HR
& 640$\times$820
& 720$\times$960
& No
& 76.81
& \textbf{86.95}
& \textbf{81.88} \\

\bottomrule
\end{tabular}
\vspace{0.5\baselineskip}
\end{table}

Overall, these findings suggest that the effectiveness of RL for embodied world models depends not only on algorithmic design, but also on optimization stability (KL and clipping), distribution alignment (resolution and rollout configuration), and exploration control (CFG). To improve reproducibility,  Appendix~\ref{sec:training_configurations} summarizes the final training settings, which correspond to the empirically stable configurations consistently used across both Kairos3.0-Robot and Cosmos-Predict2.5 experiments.

\par\endgroup

\begingroup\color{black}\colorlet{themecolor}{black}\hypersetup{linkcolor=black,citecolor=black,urlcolor=black}
\captionsetup{font={color=black},labelfont={color=black}}

\section{Conclusion and Limitations}
We present Reward as an Agent for evidence-based task verification and DynDiff-GRPO for exploration in dynamically salient regions. Experiments show benchmark gains across three world models; complementary offline evaluations show more reasonable rewards and fewer hacking/control errors.

\textbf{Limitations.} Visual and tool-based evidence can misinterpret interactions or occluded outcomes. Automatic judging on development-exposed data may overestimate generalization, particularly with a same-family judge. Independent human evaluation, matched online RL trials, and serving-cost measurements remain necessary; video benchmarks alone do not establish closed-loop robot-control gains.

\par\endgroup

\newpage
\bibliographystyle{unsrt}
\begingroup\color{black}\colorlet{themecolor}{black}\hypersetup{linkcolor=black,citecolor=black,urlcolor=black}
\bibliography{main}
\endgroup

\newpage
\appendix
\section{Appendix: Detailed Experimental Cases of Reward Hacking in Current Mainstream Reward Systems}
\label{sec:appendix_case}
This appendix presents the detailed experimental contents of the four reward hacking cases summarized in the main text, including the prompts for embodied video generation in each scenario and the corresponding scores of different reward types.

\textbf{case (a)}
\begin{jsonbox}
\begin{lstlisting}
prompt:
[robot_arm_brand:eve_robot][robot_arm_type:Dual-Arm][camera_position:head][scene_bg:office][position:in front of a desk][object:red can, silver can][task:grasped, grasped] Caption:This video is recorded using a eve_robot robotic arm (Dual-Arm) equipped with a normal camera at the head view.  The robot, equipped with dual arms, begins by simultaneously grasping two small cylindrical metal cans - a red can and a silver can - both initially positioned near the center of a red tray. The left arm moves to grasp the red can, fully closing its gripper for a stable hold, then lifts it slightly and relocates it to the left side of the tray. Concurrently, the right arm grasps the silver can, also using a fully closed gripper, lifts it slightly, and moves it to the right side of the tray while the left arm holds the red can. The two arms operate independently and synchronously throughout the task, successfully separating the cans to opposite sides of the tray without any visible abnormalities.[SCENE]: The robot operates in an office environment with bright and uniform lighting. The scene is centered around a red tray with a vertical blue stripe on its surface, placed on the floor directly in front of the robot. Behind the tray stands a white humanoid robot with a black and white body, observing the task. In the background, office furniture including tables, stools, and chairs are visible, arranged in a spacious layout. The overall setting is clean and organized, with no clutter obstructing the robot's workspace.

score:
VideoAlign
'VQ': 0.7905303239822388 (0-1)
'MQ': 0.5563995838165283 (0-1)
'TA': 0.351311057806015 (0-1)
\end{lstlisting}
\end{jsonbox}

\textbf{case (b)}
\begin{jsonbox}
\begin{lstlisting}
prompt:
This video is recorded using a Agibot-Genie-1 robotic arm (Dual-Arm) equipped with a normal camera at the head view. The robot's left arm is actively grasping an olive green, corduroy fabric sofa cushion that is resting on the left arm of the sofa. The gripper is fully closed to maintain a stable hold, and the arm is making slight adjustments to the cushion's position while keeping it securely in grasp. Meanwhile, the right arm remains stationary and does not interact with any objects. The task objective is to grasp the sofa cushion, which the left arm has successfully accomplished without any abnormal situations observed. The robot is positioned in an indoor environment with soft ambient lighting, and the scene includes an L-shaped olive green corduroy sofa occupying most of the frame, with a light brown curtain hanging vertically on the left side behind the sofa.The robot operates in an indoor setting, likely a living room, under soft ambient lighting. The primary interactive object, the olive green corduroy sofa cushion, is located on the left armrest of a large L-shaped sofa made of the same fabric. To the left of the sofa, a light brown curtain hangs vertically, partially visible behind the sofa. The robot's dual arms are positioned above the sofa, with the left arm engaged in manipulation and the right arm idle. The overall layout suggests a typical residential space with furniture arranged for comfort and functionality.

score:
\textcolor{gatingblue}{UnifiedReward-Flex}
Alignment Score (1-5): 2.756700038909912 (0-5)
Physics Score (1-5): 3.469099998474121 (0-5)
Style Score (1-5): 2.708400011062622 (0-5)
\end{lstlisting}
\end{jsonbox}

\textbf{case (c)}
\begin{jsonbox}
\begin{lstlisting}
prompt:
This video is recorded using a eve_robot robotic arm (Dual-Arm) equipped with a normal camera at the head view.The robot begins by grasping a small green rectangular object located on the wooden table near the center. Its right arm moves to pick up the object, closing its gripper fully for a stable grasp, then lifts it slightly before moving it toward the light wooden rectangular tray positioned in the center of the table. Meanwhile, the left arm remains stationary, not interacting with any objects. The task objective is to move the green object toward the tray, which the robot successfully accomplishes without any abnormal situations. The tray already holds a yellow circular object, an orange rectangular object, and a black rectangular object, which remain undisturbed during the action.[SCENE]: The robot operates in a manufacturing workshop, illuminated by natural light streaming through a large window positioned behind the table. The scene is centered around a wooden table with a light wooden rectangular tray placed in the middle. On the tray are a yellow circular object, an orange rectangular object, and a black rectangular object. In the background, various workshop equipment and tools are visible, arranged around the workspace, contributing to the industrial ambiance of the environment.

score:
VideoAlign
'VQ': 0.8031737804412842 (0-1)
'MQ': 0.382602721452713 (0-1)
'TA': 0.7577943801879883 (0-1)
\end{lstlisting}
\end{jsonbox}

\textbf{case (d)}
\begin{jsonbox}
\begin{lstlisting}
prompt: 
This video is recorded using a Agibot-Genie-1 robotic arm (Dual-Arm) equipped with a normal camera at the head view. The robot begins by coordinating its dual arms to open a brown cardboard tissue box positioned at the center of a white table. The left arm moves forward and grasps the edge of the lid, applying force to lift it, while the right arm simultaneously approaches from the opposite side, gripping the lid to assist in the opening. Both arms work in tandem, applying pressure to the lid, resulting in it being partially lifted open. The robot's actions are synchronized, and no abnormal situations are observed during this attempt. In the background, a white clipboard with a blue border rests on the left side of the table, and a blue tissue box with partially pulled-out tissue paper sits in the bottom right corner, remaining untouched throughout the process.The robot operates in a clean, well-lit indoor environment, likely a laboratory or demonstration workspace. The scene is dominated by a white table that serves as the primary workspace. On the table, the brown cardboard tissue box with a blue tissue compartment and multiple pen/pencil slots is centered, directly in front of the robot. To the left, a white clipboard with a blue border and grid paper is positioned behind the robot's left arm. In the bottom right corner, a blue tissue box with partially pulled-out tissue paper is visible. The overall setting is minimalistic, with no clutter, allowing clear visibility of the robot's manipulation task.

score:
\textcolor{gatingblue}{UnifiedReward-Flex}
Alignment Score (1-5): 1 (0-5)
Physics Score (1-5): 3 (0-5)
Style Score (1-5): 2 (0-5)
\end{lstlisting}
\end{jsonbox}

\clearpage
\begingroup\color{black}
\colorlet{themecolor}{black}
\hypersetup{linkcolor=black,citecolor=black,urlcolor=black}
\captionsetup{font={color=black},labelfont={color=black}}
\section{Appendix: Reward Agent Demonstrations}
\label{sec:appendix_case2}
This appendix illustrates how the Reward Agent converts an instruction and a generated video into an evidence-grounded reward. The seven examples cover incomplete manipulation, unsupported task execution, and successful sweeping. They use the full configuration in Sec.~\ref{sec:reward_design}, with the API-based reward model specified in Sec.~\ref{sec:reward_analysis}. The demonstrations use 32 initial frames, adaptive visual verification, temperature 0, and a 4,096-token output budget. These qualitative examples explain individual decisions; they do not constitute an additional accuracy evaluation.

\textbf{Evidence collection and verification.}
The Agent first separates task-blind observation from instruction-derived requirements, so that the requested outcome does not determine what the observer reports seeing. Joint assessment then checks the frozen requirements against the available evidence. Adaptive frame inspection and spatial crops revisit ambiguous events, while CoTracker3 supplies image-plane motion evidence and WMReward supplies latent prediction surprise. Tool-assisted reflection integrates these signals into the assessment. A requirement-scope audit checks whether the evidence and conditions used for each judgment belong to that requirement, invoking image-backed repair when necessary. The tools support verification; neither produces an independent task-reward term.

\textbf{Reward construction and interpretation.}
The reducer combines the reviewed task, physical, and visual assessments into a diagnostic score $R_{\mathrm{diag}}$. The final rules retain supported progress and assign zero to established failure or unverifiable essential video content. Training eligibility is checked separately: a returned zero is usable only when the result is marked eligible. Table~\ref{tab:qwen_demos} reports both outputs. All seven examples completed the two required tools; six produced eligible rewards, while Demo 05 retained an unresolved requirement and was marked ineligible. WMReward values are raw latent surprise, not task-success probabilities. The verdicts and scores below are the Agent's recorded judgments, rather than independently verified annotations.

\begin{table}[H]
\centering\small\color{black}
\setlength{\tabcolsep}{4pt}
\caption{Recorded outputs for seven full-Agent demonstrations. $R_{\mathrm{diag}}$ is the diagnostic score before final rules; $R$ is the returned reward. Eligible indicates whether the result can be used for training. An ineligible returned zero is an API placeholder and corresponds to $\bot$ in Eq.~\ref{eq:current_reward_gate}. WMReward reports raw latent surprise.}
\label{tab:qwen_demos}
\begin{tabular}{cllrrrc}
\toprule
Demo & Task & Verdict & WMReward & $R_{\mathrm{diag}}$ & $R$ & Eligible \\
\midrule
01 & Cloth manipulation & Mostly complete & 0.4671 & 0.860 & 0.860 & Yes \\
02 & Drawer opening & Partial & 0.4551 & 0.595 & 0.595 & Yes \\
03 & Basket manipulation & Failed & 0.4813 & 0.08125 & 0.000 & Yes \\
04 & Cube placement & Failed & 0.4089 & 0.280 & 0.000 & Yes \\
05 & Box relocation & Failed & 0.4281 & 0.280 & 0.000 & No \\
06 & Sweep carton and peel & Complete & 0.4645 & 1.000 & 1.000 & Yes \\
07 & Sweep several items & Complete & 0.4653 & 1.000 & 1.000 & Yes \\
\bottomrule
\end{tabular}
\end{table}

Each example below links the assessed visual evidence to the resulting reward. The four displayed frames summarize the video chronologically; the Agent uses the larger evidence set described above. Demos 01--02 illustrate graded progress, Demos 03--05 distinguish uncertainty, failure, and eligibility, and Demos 06--07 illustrate retained credit for supported completion.

\clearpage
\textbf{Demo 01: Cloth manipulation.}\par\smallskip
\textit{Task.} Approach and grasp the green cloth, lift it, and move it toward the back of the countertop.
\begin{figure}[H]
\centering
\includegraphics[width=.24\linewidth]{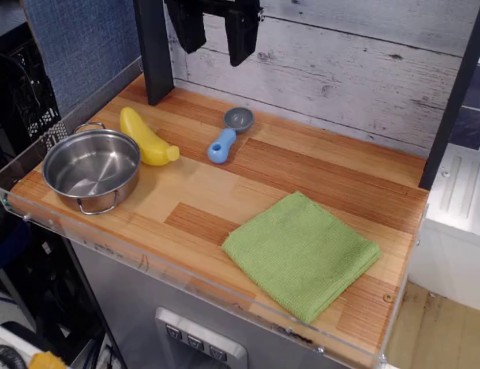}\hfill
\includegraphics[width=.24\linewidth]{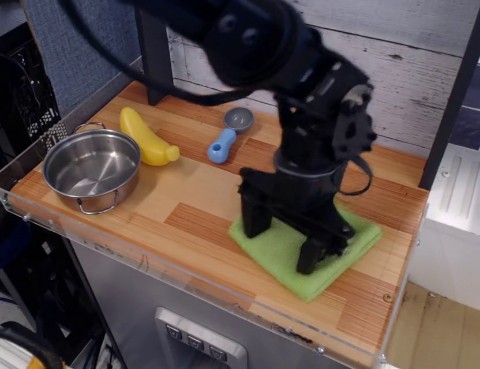}\hfill
\includegraphics[width=.24\linewidth]{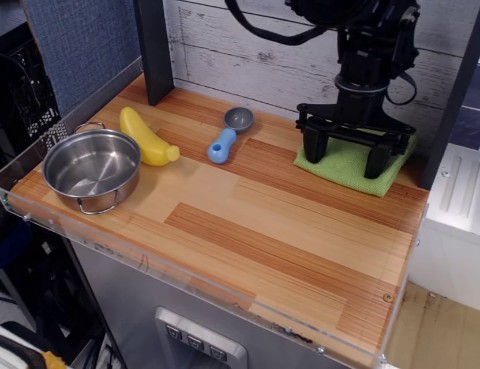}\hfill
\includegraphics[width=.24\linewidth]{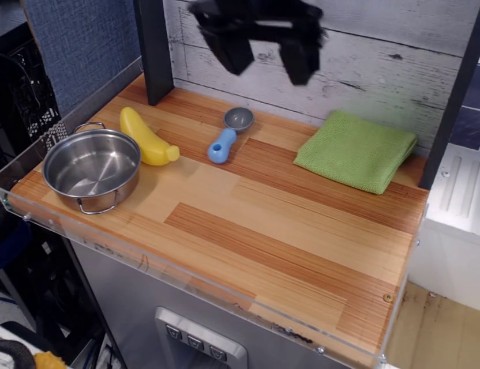}
\caption{Cloth manipulation, source frames 0, 26, 53, 80.}
\label{fig:qwen_demo01}
\end{figure}
\textit{Evidence and reward.} The Agent credits movement of the cloth toward the back of the countertop, but the evidence does not fully establish lifting clear of the surface. It therefore records mostly complete execution, with $R_{\mathrm{diag}}=R=0.860$. This example shows how the assessment can retain credit for observed displacement while leaving a manipulation requirement incompletely supported. The high score reflects the Agent's interpretation of the clip, not independent confirmation of grasping or lifting.
\par\medskip
\textbf{Demo 02: Refrigerator drawer opening.}\par\smallskip
\textit{Task.} Keep one arm stationary while the other fully opens the drawer and releases its handle.
\begin{figure}[H]
\centering
\includegraphics[width=.24\linewidth]{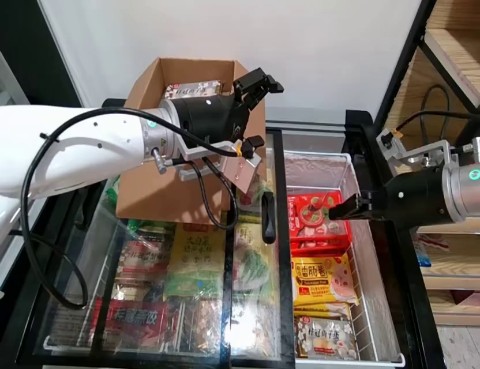}\hfill
\includegraphics[width=.24\linewidth]{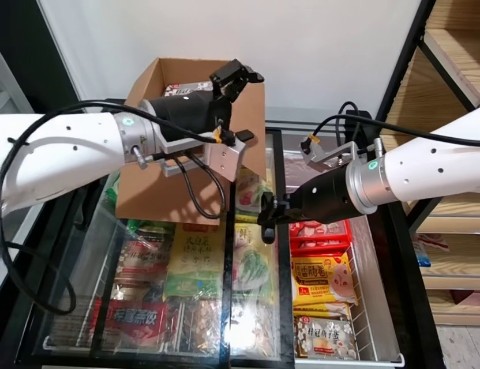}\hfill
\includegraphics[width=.24\linewidth]{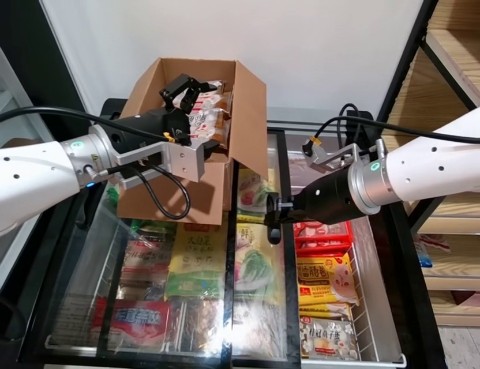}\hfill
\includegraphics[width=.24\linewidth]{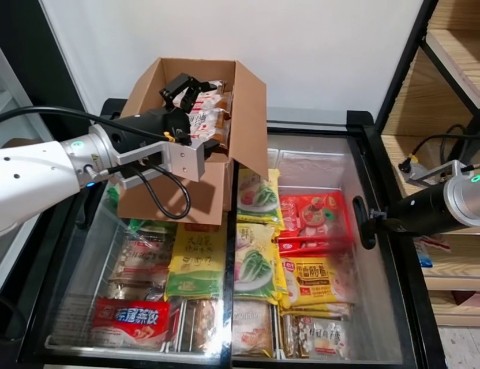}
\caption{Refrigerator drawer opening, source frames 0, 26, 53, 80.}
\label{fig:qwen_demo02}
\end{figure}
\textit{Evidence and reward.} The Agent identifies drawer-opening progress, while reporting movement of the arm required to remain stationary and continued contact with the handle at the end. It assigns partial completion and $R_{\mathrm{diag}}=R=0.595$. The example illustrates requirement-specific assessment: progress on opening the drawer receives credit, but does not establish compliance with the stationary-arm and release conditions. The scope audit accepts the report without repair.
\medskip

\clearpage
\textbf{Demo 03: Can placement and basket-handle grasping.}\par\smallskip
\textit{Task.} Lower the red can into the basket with the right arm, then grasp the basket handle with the left arm.
\begin{figure}[H]
\centering
\includegraphics[width=.24\linewidth]{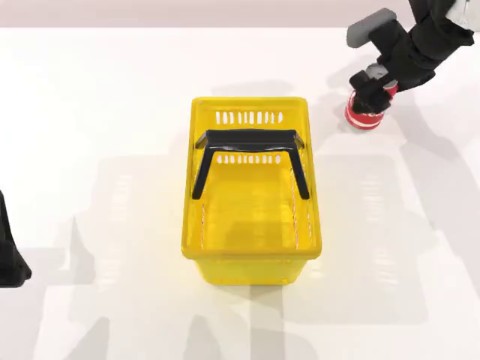}\hfill
\includegraphics[width=.24\linewidth]{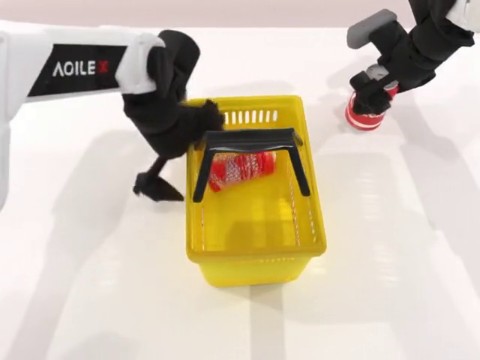}\hfill
\includegraphics[width=.24\linewidth]{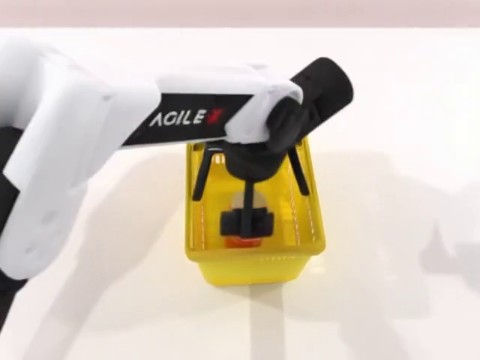}\hfill
\includegraphics[width=.24\linewidth]{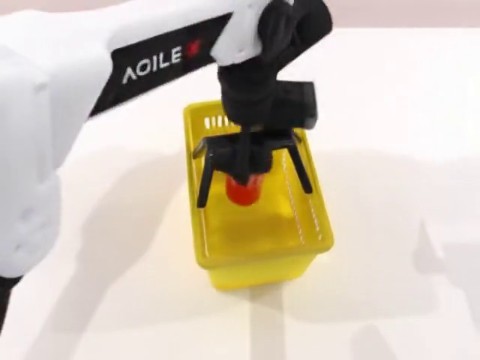}
\caption{Can placement and basket-handle grasping, source frames 0, 40, 80, 120.}
\label{fig:qwen_demo03}
\end{figure}
\textit{Evidence and reward.} The Agent does not establish the instructed right-arm can placement followed by left-arm basket-handle grasping. Its report records a failed verdict, but lacks sufficient confidence to treat that verdict as confirmed failure. The video-uncertainty rule therefore maps $R_{\mathrm{diag}}=0.08125$ to an eligible $R=0$. Here, zero denotes withheld credit for unsupported execution; it does not establish the Agent's failure interpretation as ground truth.
\par\medskip
\textbf{Demo 04: Green cube placement.}\par\smallskip
\textit{Task.} Grasp and place the green cube with the right arm while keeping the left arm stationary.
\begin{figure}[H]
\centering
\includegraphics[width=.24\linewidth]{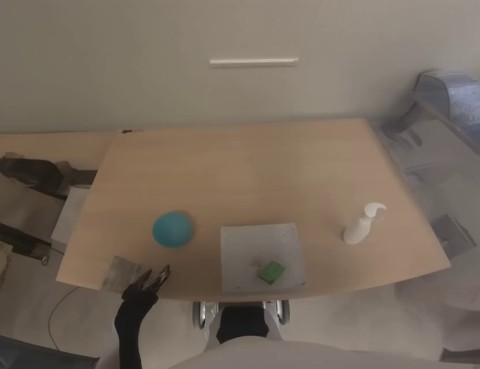}\hfill
\includegraphics[width=.24\linewidth]{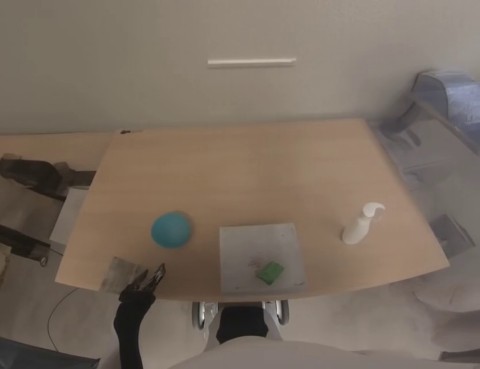}\hfill
\includegraphics[width=.24\linewidth]{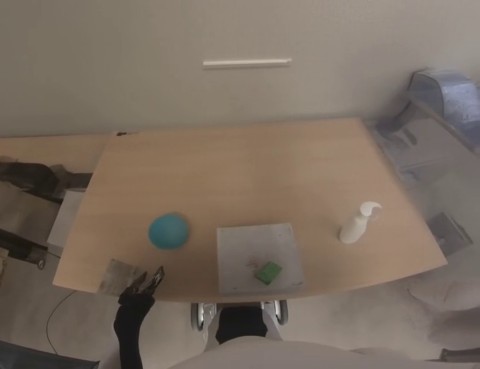}\hfill
\includegraphics[width=.24\linewidth]{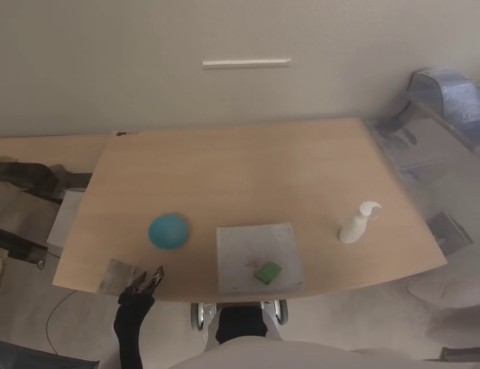}
\caption{Green cube placement, source frames 0, 26, 53, 80.}
\label{fig:qwen_demo04}
\end{figure}
\textit{Evidence and reward.} The Agent finds no visible grasp or placement of the green cube. Keeping the other arm stationary satisfies only a supporting condition and cannot substitute for the requested manipulation. Although the combined diagnostic score is $R_{\mathrm{diag}}=0.280$, the failure rule yields an eligible $R=0$. This case illustrates the intended protection against positive reward from favorable physical or visual assessments when essential task execution is absent.
\medskip

\clearpage
\textbf{Demo 05: Box relocation.}\par\smallskip
\textit{Task.} Grasp, lift, and move the box to the right while the other arm remains stationary.
\begin{figure}[H]
\centering
\includegraphics[width=.24\linewidth]{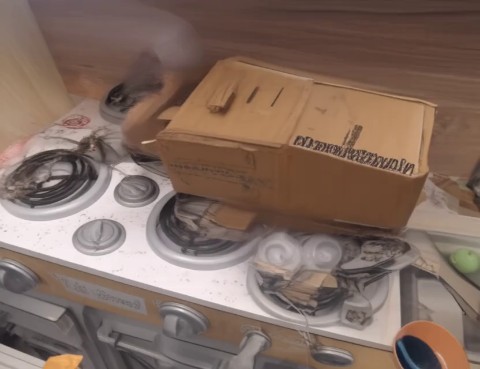}\hfill
\includegraphics[width=.24\linewidth]{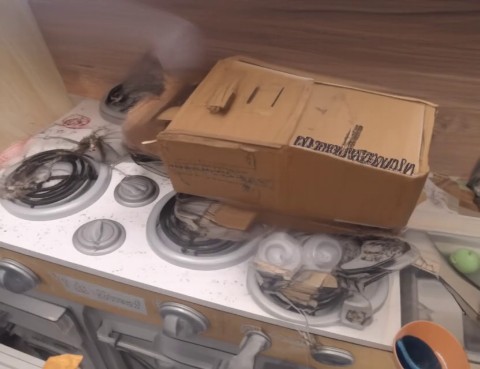}\hfill
\includegraphics[width=.24\linewidth]{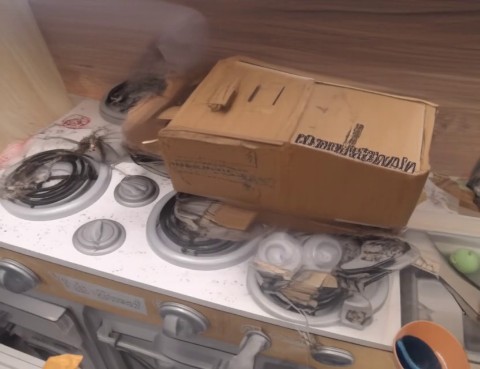}\hfill
\includegraphics[width=.24\linewidth]{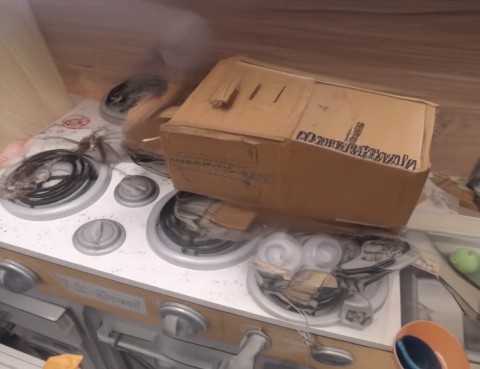}
\caption{Box relocation, source frames 0, 26, 53, 80.}
\label{fig:qwen_demo05}
\end{figure}
\textit{Evidence and reward.} The Agent reports no visible grasp, lift, or rightward relocation of the box, while the requirement concerning the other arm remains unresolved. It returns $R_{\mathrm{diag}}=0.280$ and $R=0$, but marks the result ineligible for training. Unlike the eligible zeros in Demos 03--04, this output cannot be used as a training reward without resolving or masking the report. The case exposes a limitation of the verification workflow: completing both tools does not ensure that every requirement is resolved.
\par\medskip
\textbf{Demo 06: Sweeping a carton and peel.}\par\smallskip
\textit{Task.} Sweep the trash into the dustpan using a broom.
\begin{figure}[H]
\centering
\includegraphics[width=.24\linewidth]{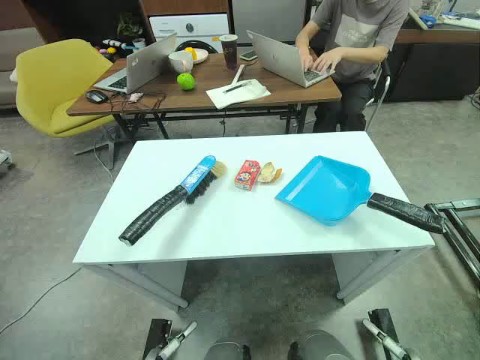}\hfill
\includegraphics[width=.24\linewidth]{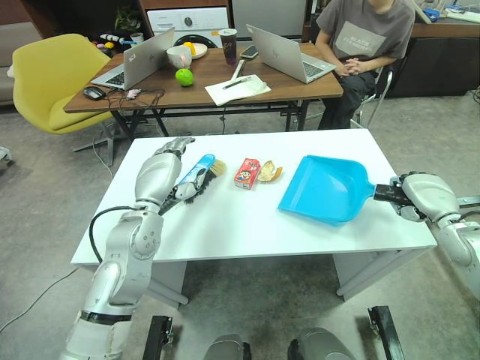}\hfill
\includegraphics[width=.24\linewidth]{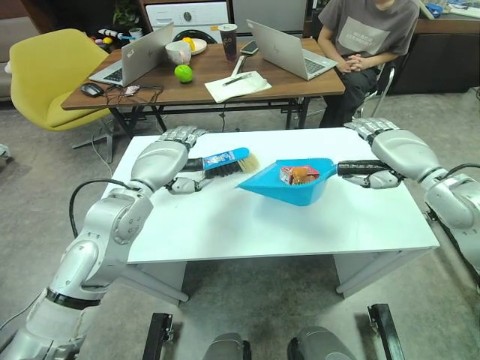}\hfill
\includegraphics[width=.24\linewidth]{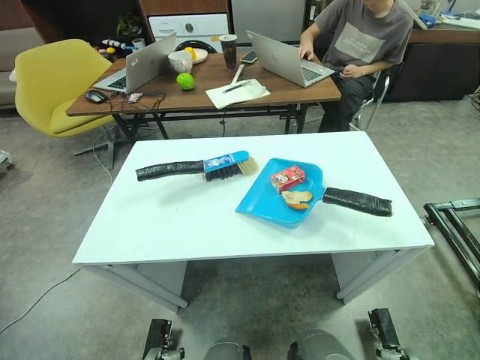}
\caption{Sweeping a carton and peel, source frames 0, 372, 745, 1118.}
\label{fig:qwen_demo06}
\end{figure}
\textit{Evidence and reward.} The Agent identifies broom use and a final state in which the carton and peel are inside the dustpan. It assigns complete execution and $R_{\mathrm{diag}}=R=1$. Although individual views do not expose every intermediate motion, the observed interaction and final placement support its completion judgment. Assessment remains tied to the instruction, without adding requirements for tidiness or motion style.
\medskip

\clearpage
\textbf{Demo 07: Sweeping several pieces of litter.}\par\smallskip
\textit{Task.} Sweep the trash into the dustpan using a broom.
\begin{figure}[H]
\centering
\includegraphics[width=.24\linewidth]{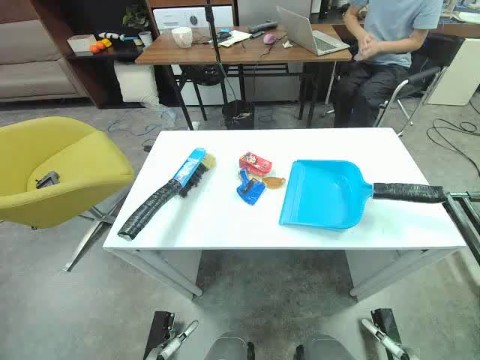}\hfill
\includegraphics[width=.24\linewidth]{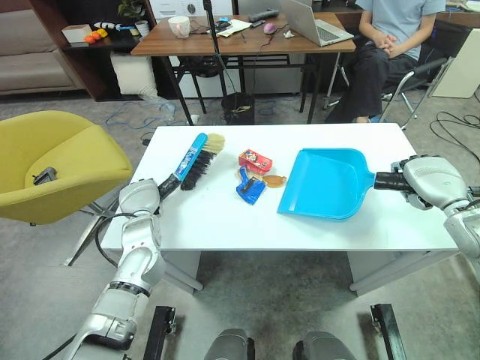}\hfill
\includegraphics[width=.24\linewidth]{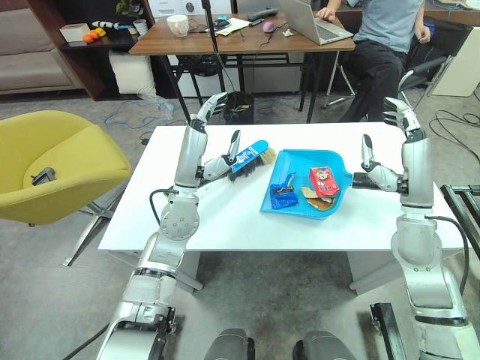}\hfill
\includegraphics[width=.24\linewidth]{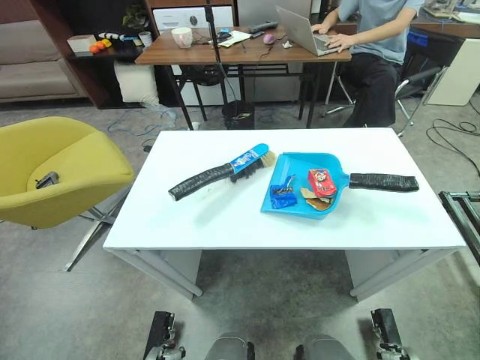}
\caption{Sweeping several pieces of litter, source frames 0, 419, 839, 1259.}
\label{fig:qwen_demo07}
\end{figure}
\textit{Evidence and reward.} The Agent reports that the litter is swept into the dustpan and remains there in later and final frames. Tool-assisted reflection and scope checking retain the complete verdict, yielding $R_{\mathrm{diag}}=R=1$. Together with Demo 06, this case illustrates preservation of positive credit when the required interaction and resulting state are supported by the video.
\medskip
\clearpage

\endgroup
\begingroup\color{black}\colorlet{themecolor}{black}\hypersetup{linkcolor=black,citecolor=black,urlcolor=black}\reviewchangesfalse
\captionsetup{font={color=black},labelfont={color=black}}
\section{Appendix: Reward Service Deployment}
\label{sec:reward_service}
\textbf{API-based deployment.}
The Doubao branch of the offline reward evaluation and the recorded demonstrations use Doubao 2.1 Pro through its API. The evaluation also includes Qwen3.8-Flash-Next, as reported in Sec.~\ref{sec:reward_analysis}. The full service configuration was checked against repository revision \texttt{199fc6c}. Model inference is remote, while WMReward runs in one resident worker and CoTracker3 runs in the Agent process; the two tools share a local GPU. No local base-model server is required; Table~\ref{tab:current_reward_config} summarizes the service settings.

\begin{table}[H]
\centering\small\color{black}
\caption{Settings of the API-based full reward service.}
\label{tab:current_reward_config}
\begin{tabular}{lp{.63\linewidth}}
\toprule
Setting & Effective value \\
\midrule
Base model & Doubao 2.1 Pro via API \\
Decoding & Temperature 0; thinking disabled; maximum 4,096 output tokens per call \\
Initial evidence & 32 uniformly sampled original frames \\
Adaptive inspection & Eight planning frames; one or two enlarged regions; CoTracker3 motion overlays \\
Verification & All frames for clips of at most 81 frames; bounded neighboring frames for longer clips \\
External tools & WMReward with required reflection; CoTracker3 \texttt{scaled\_offline} \\
Requirement checks & Joint scope audit; at most three image-backed repair rounds \\
Execution & Service concurrency limit 8; one resident WMReward worker \\
Reuse / archival & In-process task-contract cache; source-matched motion cache; input archival enabled \\
Gating & Optional process policy off; native failure and video-uncertainty gates on \\
Fast mode & Disabled \\
\bottomrule
\end{tabular}
\end{table}

Adaptive crops, motion evidence, WMReward reflection, and scope repair remain enabled throughout the full profile. Although the environment allows 8,192 output tokens, the pipeline caps each call at 4,096. Task contracts and source-matched motion evidence can be cached; the ablation evaluation uses fresh model calls except for identical within-run stage inputs. API, output-validation, and required-tool failures follow bounded retry policies. Training clients must check the returned eligibility flag and retry or exclude ineligible results. Tool outputs retain source and checkpoint fingerprints; the recorded WMReward and CoTracker3 revisions are \texttt{f2af5373} and \texttt{82e02e80}, respectively.

\textbf{Alternative local-model deployment.}
The same Agent can use Qwen3.8-Flash-Next through a local vLLM~\cite{kwon2023efficient} OpenAI-compatible endpoint. Switching the provider, API base, and served model name preserves the full Agent's evidence tools, verification stages, and reward rules. The local deployment uses four-way tensor parallelism, BF16 inference, CPU offloading of PLE, a 131,072-token context limit, and up to 128 images per request. Both backbones use the same evidence tools, verification stages, and reward rules.

\textbf{Training deployment.}
The world-model training experiments in Table~\ref{tab:paibench_robot} use Doubao 2.1 Pro through its API as the Reward Agent backbone. The same Agent workflow also supports the open-source Qwen Flash backbone through local vLLM serving. The offline Agent evaluation compares Doubao 2.1 Pro and Qwen3.8-Flash-Next; the recorded demonstrations use Doubao 2.1 Pro via API. These experiments assess world-model training performance and reward quality, respectively; their model backends and compute settings are reported separately.

\section{Appendix: DynDiff-GRPO Implementation}
\label{sec:DynDiffgrpo_code}
We summarize the implementation of DynDiff-GRPO rollout in Algorithm~\ref{alg:dyn_diff_final}. The method consists of three stages: dynamic-aware mask construction, hybrid ODE/SDE sampling, and likelihood evaluation with spatial variance. The algorithm describes stochastic steps; a zero-noise terminal step returns the deterministic mean without a Gaussian likelihood. 

\begingroup\footnotesize
\begin{algorithm}[H]
\caption{DynDiff-GRPO: \textcolor{black}{Dynamic-Aware} Rollout}
\label{alg:dyn_diff_final}
\begin{algorithmic}[1]

\Require latent state $x_t$, condition $c$, velocity model $v_\theta$, noise levels $\sigma_t,\sigma_{t^-}>0$, base ratio $0<r_{\text{base}}\leq1$, percentile $\tau$
\Ensure sample $x_{t^-}$, log-probability $\log p(x_{t^-})$

\vspace{0.3em}
\State \textbf{Stage 1: Dynamic-aware mask construction}

\State $\hat{x}_0 \gets x_t - \sigma_t v_\theta(x_t,t,c)$
\State $\Delta \hat{x}_0 \gets \mathrm{TemporalDiff}(\hat{x}_0)$
\State $R \gets \Delta \hat{x}_0 - \mathrm{Mean}_{h,w}(\Delta \hat{x}_0)$
\State $M \gets \|R\|_2$

\State $M \gets \mathrm{Normalize}(M)$
\State $B \gets \mathbb{I}[M > \mathrm{Quantile}(M,\tau)]$

\State $M_t \gets r_{\text{base}} + (1-r_{\text{base}})\cdot(0.5M + 0.5B)$

\vspace{0.5em}
\State \textbf{Stage 2: hybrid ODE/SDE sampling}

\State $\hat{\epsilon}_t \gets x_t + (1-\sigma_t)v_\theta(x_t,t,c)$
\State $\sigma_{t^-}^{\text{noise}} \gets \sigma_{t^-} \odot M_t$
\State $z \sim \mathcal{N}(0,I)$

\State $a \gets \hat{x}_0 (1-\sigma_{t^-})$
\State $b \gets \hat{\epsilon}_t \sqrt{\sigma_{t^-}^2 - (\sigma_{t^-}^{\text{noise}})^2}$
\State $u_{\mathrm{noise}} \gets \sigma_{t^-}^{\text{noise}} \odot z$

\State $x_{t^-} \gets a + b + u_{\mathrm{noise}}$

\vspace{0.5em}
\State \textbf{Stage 3: Likelihood evaluation}

\State $\mu \gets a + b$
\State $\sigma \gets \sigma_{t^-} \odot M_t$

\State $d \gets x_{t^-} - \mu$
\State $\log p(x_{t^-}) \gets -\sum_j\left[\frac{d_j^2}{2\sigma_j^2} + \log \sigma_j + \log \sqrt{2\pi}\right]$

\Return $x_{t^-}, \log p(x_{t^-})$

\end{algorithmic}
\end{algorithm}
\endgroup

\par\endgroup

\section{Appendix: Training Configurations}
\label{sec:training_configurations}
As discussed in the main text, the performance of RL finetuning for embodied world models is strongly influenced by optimization stability, rollout configuration, and generation settings. To improve reproducibility and provide a complete reference for practitioners, we summarize the final training configurations adopted in our experiments in Table~\ref{tab:training_settings}. The supplementary Kairos3.1 experiments use exactly the same configuration as Kairos3.0-Robot. \textcolor{gatingblue}{These settings describe the main training runs; the resolution and CFG ablations use the exceptions stated in their respective sections. They do not describe the Agent demonstrations or the offline reward-service ablation.}

\begin{table}[H]
\centering
\footnotesize
\setlength{\tabcolsep}{6pt}
\renewcommand{\arraystretch}{1.0}
\caption{Training settings for Kairos3.0-Robot, Kairos3.1, and Cosmos-Predict2.5. Both Kairos models share the listed configuration.}
\label{tab:training_settings}
\begin{tabular}{llcc}
\toprule
\textbf{Category} & \textbf{Setting} & \textbf{Kairos3.0 / 3.1} & \textbf{Cosmos-Predict2.5} \\
\midrule

\multirow{12}{*}{Training}
& Training batch size & 4 & 4 \\
& Best of N & 8 & 8 \\
& Clip range $\delta$ & 1e-4 & 1e-4 \\
& Advantage clipping threshold & 2.5 & 2.5 \\
& Base Noise level & 0.6 & 0.6 \\
& Dynamic-intensity quantile $\tau$ & 0.55 & 0.55 \\
& CFG scale & -- & -- \\
& KL coefficient & 0.5 & 0.5 \\
& KL period steps & 2 & 2 \\
& Optimizer type & AdamW~\cite{loshchilov2019adamw} & AdamW \\
& Learning rate & 2e-6 & 2e-6 \\
& Weight decay & 0.01 & 0.001 \\

\midrule

\multirow{5}{*}{Rollout}
& Rollout batch size & 2 & 2 \\
& Prompts per iteration & 16 & 16 \\
& Videos per prompt & 16 & 16 \\
& Gradient updates per iteration & 1 & 1 \\
& Resolution & $640 \times 820$ & $640 \times 820$ \\

\bottomrule
\end{tabular}
\end{table}

\clearpage
\begingroup\colorlet{themecolor}{black}\hypersetup{linkcolor=black,citecolor=black,urlcolor=black}
\begingroup\color{black}
\section{Appendix: Evaluation Samples and Prompt}
\label{sec:evaluation_data_examples}
The evaluation in Sec.~\ref{sec:reward_analysis} uses 180 videos: 100 ordinary clips and 80 hacking/control clips. All configurations are evaluated on the same videos and instructions. Table~\ref{tab:reward_system_eval_matrix} reports reward reasonableness over all 180 clips. Table~\ref{tab:reward_hacking_suppression} uses the 80-clip subset, comprising 57 reference failures, 3 unobservable clips, and 20 clips with partial or complete progress. Its two counts measure positive rewards on the first 60 clips and zero rewards on the 20 progress clips. These automatic reference labels are separate from the reasonableness judgments and are withheld from the judge.

The examples below illustrate ordinary recordings and several challenge types. Each strip contains four chronological frames, with its instruction, timestamps, and sample ID. These illustrations describe the inputs; they do not report additional per-example scores or independent success labels. The review prompt appears in Appendix~\ref{sec:reward_judge_exact_prompt}.

\subsection{Ordinary Video Samples}
Figures~\ref{fig:eval_regular_039} and~\ref{fig:eval_regular_080} show bottle lifting and can placement from the ordinary-video subset. The evaluation considers the instructed interaction and resulting state, rather than visual appearance alone.

\begin{figure}[H]
\color{black}
\centering
\captionsetup{font={small,color=black},labelfont={color=black}}
\textbf{Instruction: Lift red-capped bottle}\par
\smallskip
\begin{minipage}{0.242\linewidth}\centering
\includegraphics[width=\linewidth]{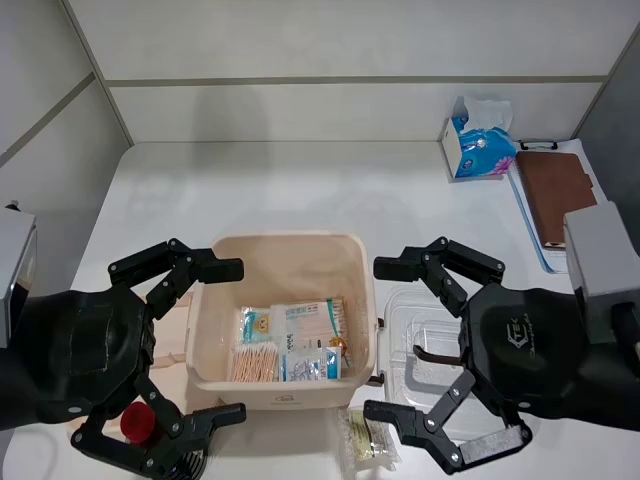}\par
{\scriptsize $t=0.00\,\mathrm{s}$}
\end{minipage}\hfill
\begin{minipage}{0.242\linewidth}\centering
\includegraphics[width=\linewidth]{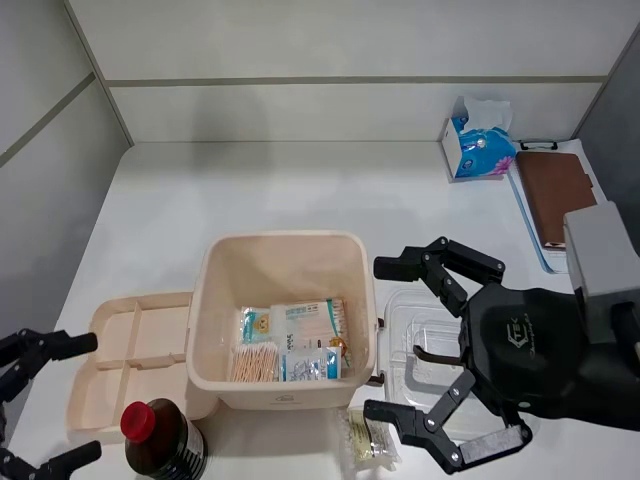}\par
{\scriptsize $t=2.00\,\mathrm{s}$}
\end{minipage}\hfill
\begin{minipage}{0.242\linewidth}\centering
\includegraphics[width=\linewidth]{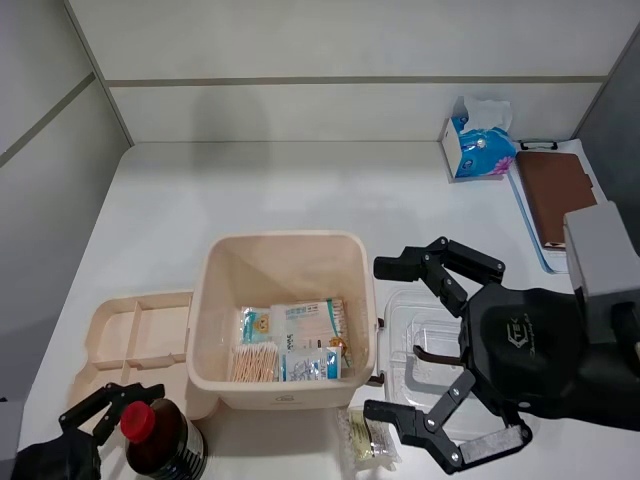}\par
{\scriptsize $t=4.00\,\mathrm{s}$}
\end{minipage}\hfill
\begin{minipage}{0.242\linewidth}\centering
\includegraphics[width=\linewidth]{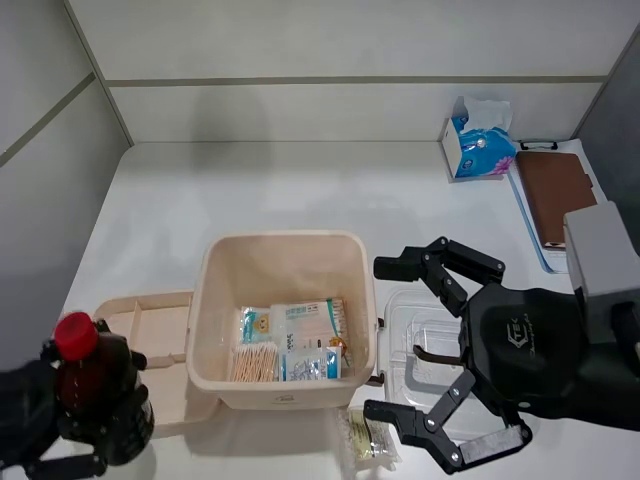}\par
{\scriptsize $t=6.00\,\mathrm{s}$}
\end{minipage}
\caption{Bottle-lifting sample. The sequence shows the gripper approaching the red-capped bottle and moving it upward. Sample: \texttt{regular\_039}.}
\label{fig:eval_regular_039}
\end{figure}

\begin{figure}[H]
\color{black}
\centering
\captionsetup{font={small,color=black},labelfont={color=black}}
\textbf{Instruction: Drop green can into cup}\par
\smallskip
\begin{minipage}{0.242\linewidth}\centering
\includegraphics[width=\linewidth]{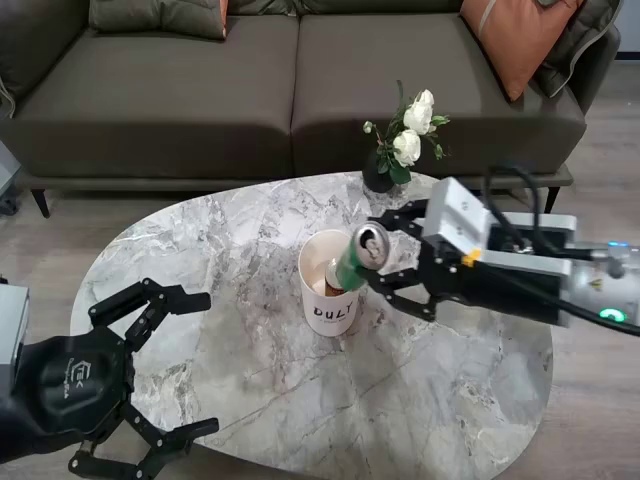}\par
{\scriptsize $t=0.00\,\mathrm{s}$}
\end{minipage}\hfill
\begin{minipage}{0.242\linewidth}\centering
\includegraphics[width=\linewidth]{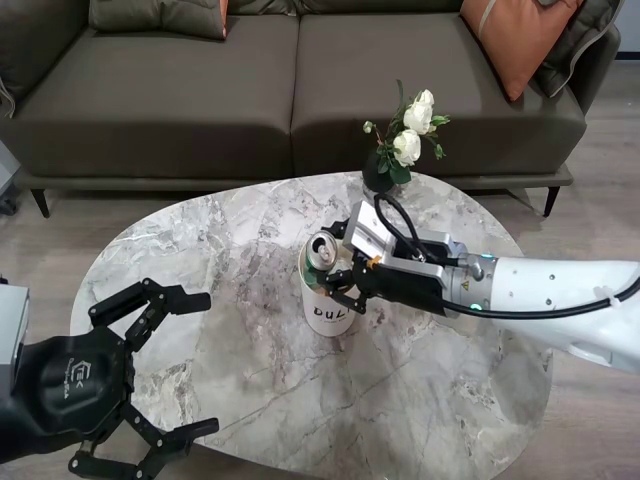}\par
{\scriptsize $t=1.62\,\mathrm{s}$}
\end{minipage}\hfill
\begin{minipage}{0.242\linewidth}\centering
\includegraphics[width=\linewidth]{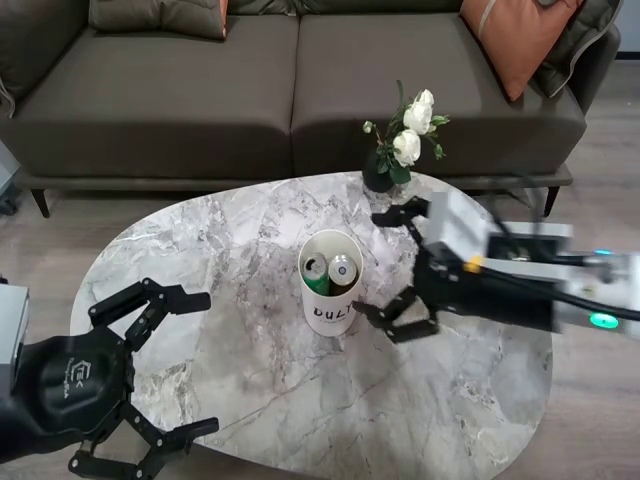}\par
{\scriptsize $t=3.19\,\mathrm{s}$}
\end{minipage}\hfill
\begin{minipage}{0.242\linewidth}\centering
\includegraphics[width=\linewidth]{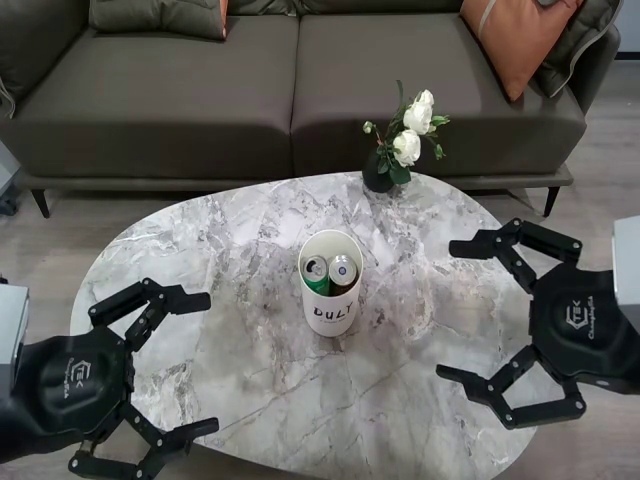}\par
{\scriptsize $t=4.81\,\mathrm{s}$}
\end{minipage}
\caption{Can-placement sample. The sequence shows the green can being placed into the cup, followed by gripper withdrawal. Sample: \texttt{regular\_080}.}
\label{fig:eval_regular_080}
\end{figure}

\clearpage
\subsection{Absent Action and Obscured Outcome}
Figures~\ref{fig:eval_hacking_001_challenge} and~\ref{fig:eval_hacking_013_challenge} illustrate challenges to task-evidence verification: preserving scene appearance without action, and obscuring the outcome needed to assess the instruction. The review criterion distinguishes lack of useful progress from insufficient essential visual evidence; either can justify zero reward.

\begin{figure}[H]
\color{black}
\centering
\captionsetup{font={small,color=black},labelfont={color=black}}
\textbf{Instruction: Re-grasp and lift garment}\par
\smallskip
\begin{minipage}{0.242\linewidth}\centering
\includegraphics[width=\linewidth]{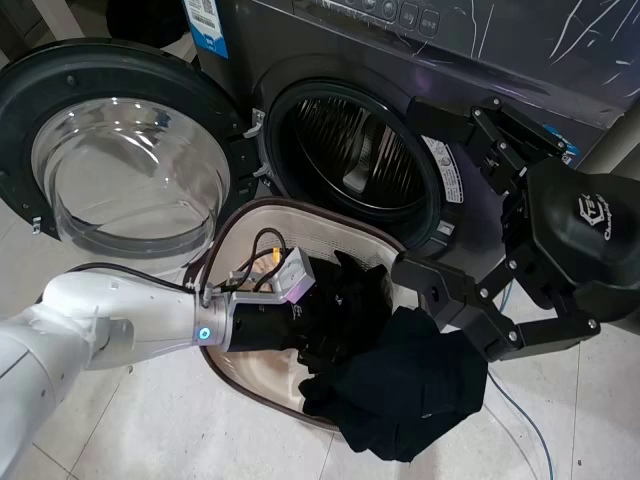}\par
{\scriptsize $t=0.00\,\mathrm{s}$}
\end{minipage}\hfill
\begin{minipage}{0.242\linewidth}\centering
\includegraphics[width=\linewidth]{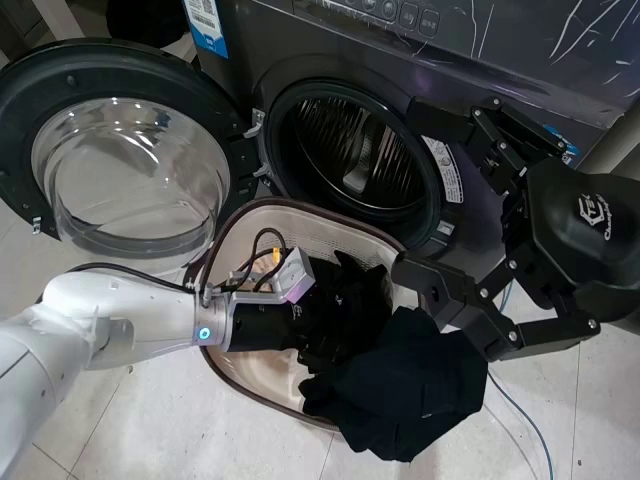}\par
{\scriptsize $t=1.31\,\mathrm{s}$}
\end{minipage}\hfill
\begin{minipage}{0.242\linewidth}\centering
\includegraphics[width=\linewidth]{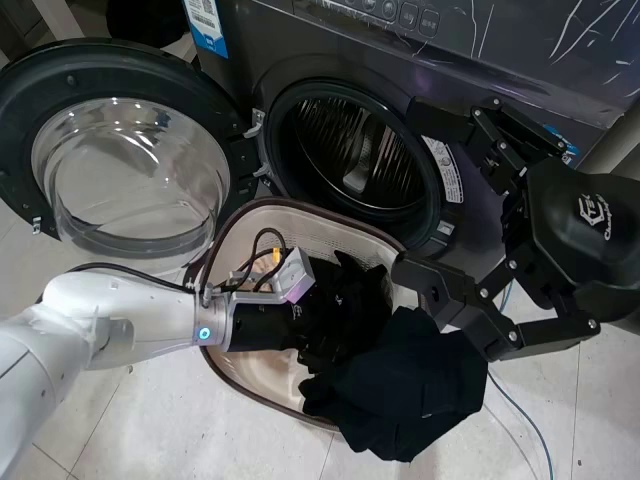}\par
{\scriptsize $t=2.69\,\mathrm{s}$}
\end{minipage}\hfill
\begin{minipage}{0.242\linewidth}\centering
\includegraphics[width=\linewidth]{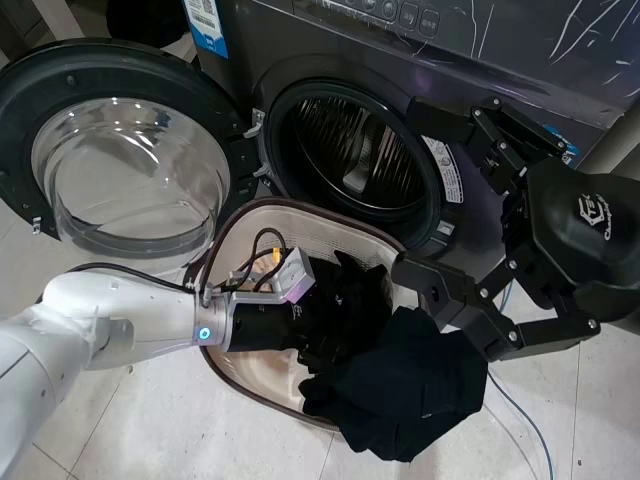}\par
{\scriptsize $t=4.00\,\mathrm{s}$}
\end{minipage}
\caption{No-action challenge. The first frame is repeated throughout the clip, preserving scene appearance while removing the requested re-grasping and lifting. Sample: \texttt{hacking\_001\_challenge}.}
\label{fig:eval_hacking_001_challenge}
\end{figure}

\begin{figure}[H]
\color{black}
\centering
\captionsetup{font={small,color=black},labelfont={color=black}}
\textbf{Instruction: Release garment edge}\par
\smallskip
\begin{minipage}{0.242\linewidth}\centering
\includegraphics[width=\linewidth]{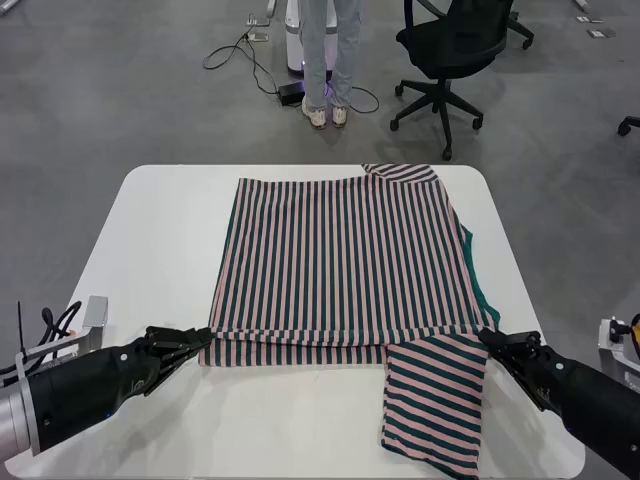}\par
{\scriptsize $t=0.00\,\mathrm{s}$}
\end{minipage}\hfill
\begin{minipage}{0.242\linewidth}\centering
\includegraphics[width=\linewidth]{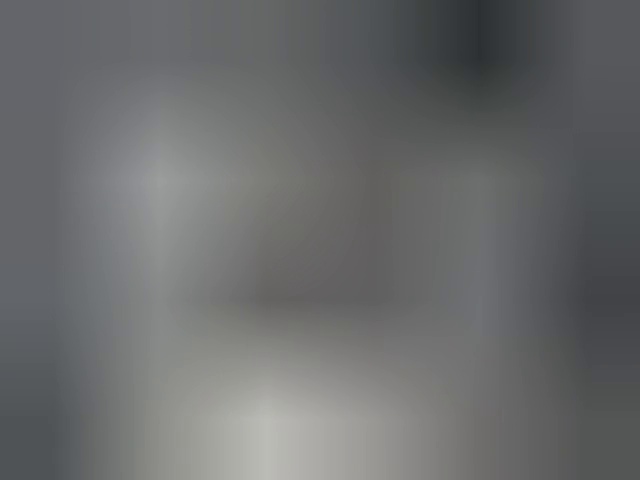}\par
{\scriptsize $t=2.00\,\mathrm{s}$}
\end{minipage}\hfill
\begin{minipage}{0.242\linewidth}\centering
\includegraphics[width=\linewidth]{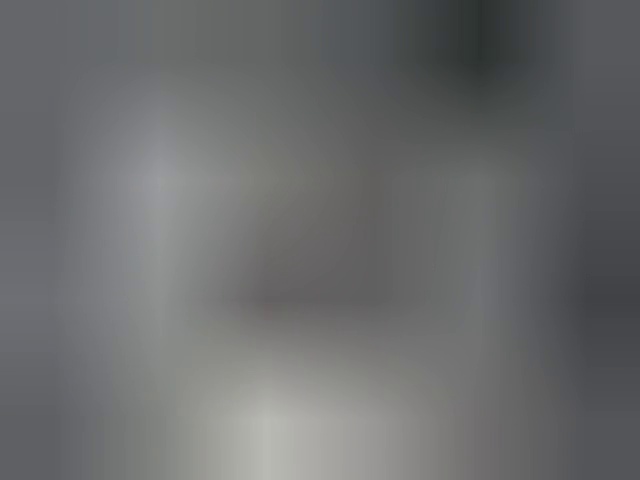}\par
{\scriptsize $t=4.00\,\mathrm{s}$}
\end{minipage}\hfill
\begin{minipage}{0.242\linewidth}\centering
\includegraphics[width=\linewidth]{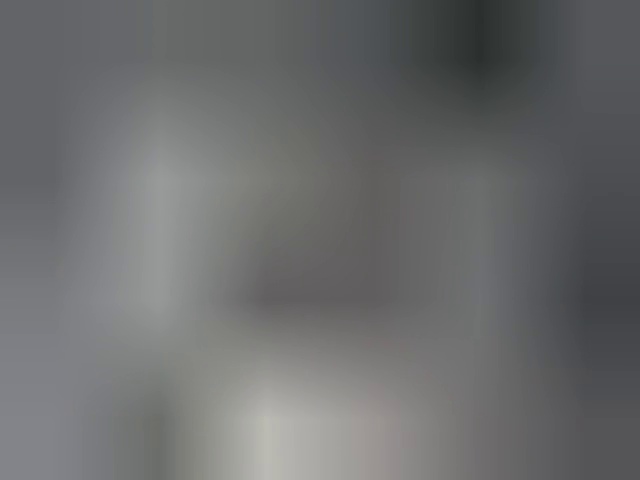}\par
{\scriptsize $t=6.00\,\mathrm{s}$}
\end{minipage}
\caption{Outcome-occlusion challenge. From frame 14 onward, each frame is reduced to $6\times4$ pixels and enlarged back to the original resolution. This edit obscures the evidence needed to determine whether the garment edge is released; an unchanged scene or plausible appearance cannot establish the requested outcome. Sample: \texttt{hacking\_013\_challenge}.}
\label{fig:eval_hacking_013_challenge}
\end{figure}

\clearpage
\subsection{Disrupted Interaction and Temporal Evidence}
Figures~\ref{fig:eval_hacking_025_challenge} and~\ref{fig:eval_hacking_031_challenge} illustrate spatial and temporal edits that challenge verification of the instructed interaction. An edit alone does not determine a failure label: the evaluator must assess what task-relevant evidence remains visible.

\begin{figure}[H]
\color{black}
\centering
\captionsetup{font={small,color=black},labelfont={color=black}}
\textbf{Instruction: Place block in tray.}\par
\smallskip
\begin{minipage}{0.242\linewidth}\centering
\includegraphics[width=\linewidth]{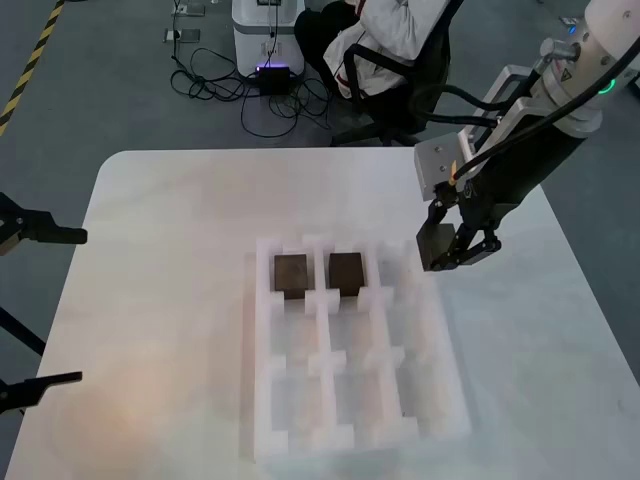}\par
{\scriptsize $t=0.00\,\mathrm{s}$}
\end{minipage}\hfill
\begin{minipage}{0.242\linewidth}\centering
\includegraphics[width=\linewidth]{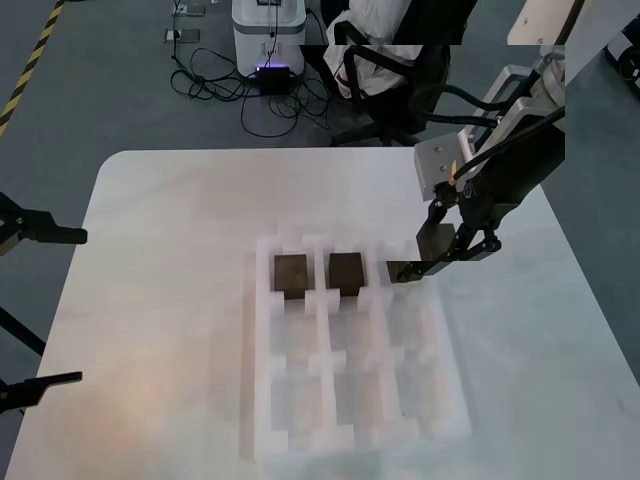}\par
{\scriptsize $t=2.00\,\mathrm{s}$}
\end{minipage}\hfill
\begin{minipage}{0.242\linewidth}\centering
\includegraphics[width=\linewidth]{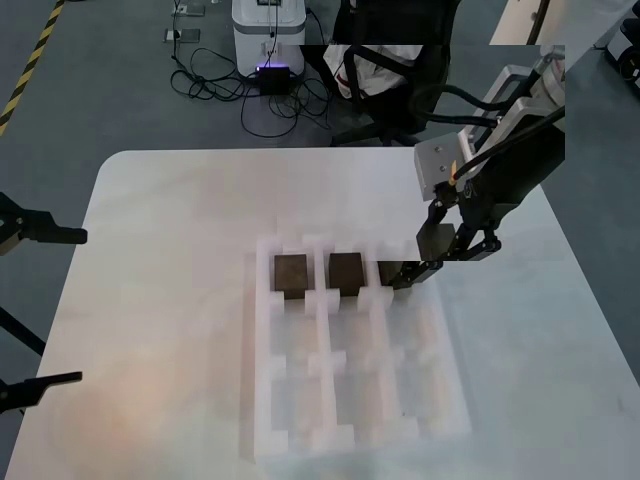}\par
{\scriptsize $t=4.00\,\mathrm{s}$}
\end{minipage}\hfill
\begin{minipage}{0.242\linewidth}\centering
\includegraphics[width=\linewidth]{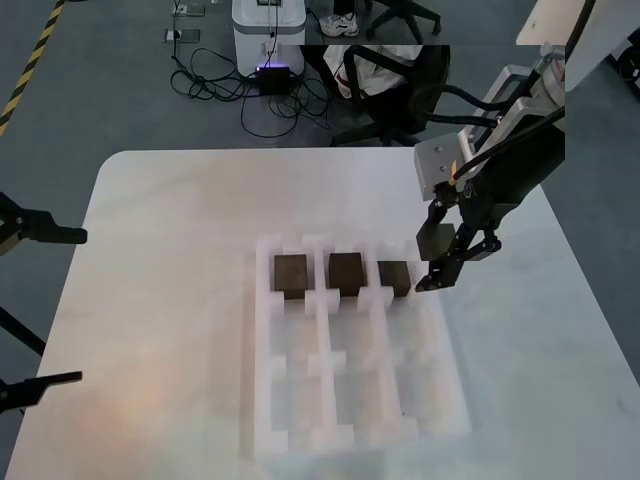}\par
{\scriptsize $t=6.00\,\mathrm{s}$}
\end{minipage}
\caption{Interaction-disruption challenge. A $288\times216$ motion-centered region at $(x,y)=(277,45)$ is replaced with its first-frame content throughout the clip. The edit interrupts visible interaction evidence while retaining surrounding appearance. Freezing a region does not by itself establish semantic object removal. Sample: \texttt{hacking\_025\_challenge}.}
\label{fig:eval_hacking_025_challenge}
\end{figure}

\begin{figure}[H]
\color{black}
\centering
\captionsetup{font={small,color=black},labelfont={color=black}}
\textbf{Instruction: Push freezer drawer shut}\par
\smallskip
\begin{minipage}{0.242\linewidth}\centering
\includegraphics[width=\linewidth]{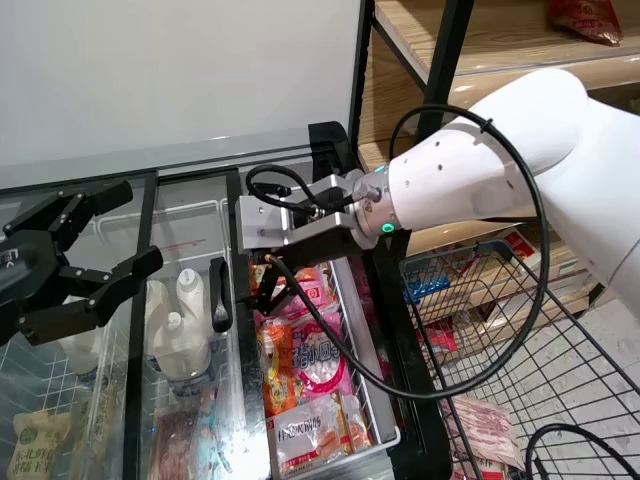}\par
{\scriptsize $t=0.00\,\mathrm{s}$}
\end{minipage}\hfill
\begin{minipage}{0.242\linewidth}\centering
\includegraphics[width=\linewidth]{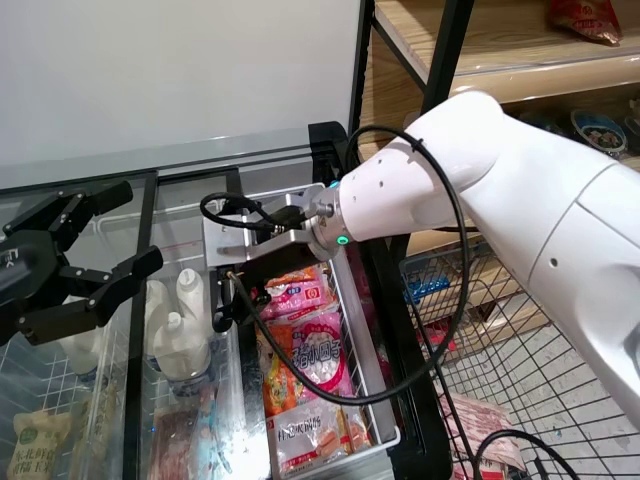}\par
{\scriptsize $t=1.69\,\mathrm{s}$}
\end{minipage}\hfill
\begin{minipage}{0.242\linewidth}\centering
\includegraphics[width=\linewidth]{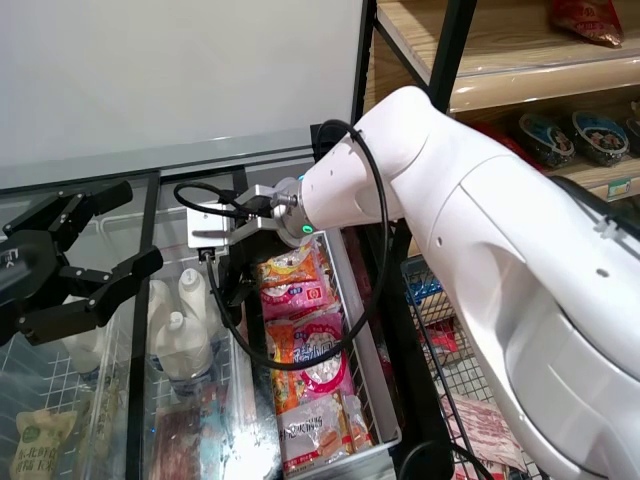}\par
{\scriptsize $t=3.31\,\mathrm{s}$}
\end{minipage}\hfill
\begin{minipage}{0.242\linewidth}\centering
\includegraphics[width=\linewidth]{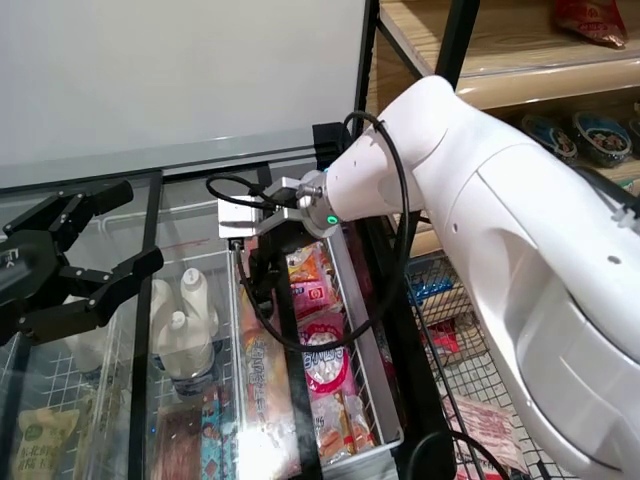}\par
{\scriptsize $t=5.00\,\mathrm{s}$}
\end{minipage}
\caption{Temporal-discontinuity candidate. Source-clip frames 24--55 are removed and the remaining frames are resampled to the original length. The removed interval can hide intermediate actions; evaluation must consider the remaining evidence for drawer closure. Temporal editing alone does not establish a physical violation. Sample: \texttt{hacking\_031\_challenge}.}
\label{fig:eval_hacking_031_challenge}
\end{figure}

\endgroup

\clearpage
\subsection{Automatic Reward Evaluation Prompt}
\label{sec:reward_judge_prompt}
\label{sec:reward_judge_exact_prompt}
\begingroup\color{black}
The judge reviews up to 64 video frames, the instruction, reward, and original scoring explanation at temperature 0. Method identities and reference labels are withheld. It returns \texttt{unreasonable}=0 for reasonable and 1 for unreasonable; Table~\ref{tab:reward_system_eval_matrix} counts reasonable judgments over 180 clips. Table~\ref{tab:reward_hacking_suppression} separately compares rewards with reference labels. The evaluation system prompt is reproduced below.

\begingroup\small\raggedright
Assess the OVERALL reasonableness of a robot-video reward. Use a broad, practical standard, equally for every anonymous scoring method. You are not auditing every sentence or computing an exact reference score. Do not aim for a particular acceptance rate or method ranking.\par\smallskip

You receive only the task instruction, sampled video frames, proposed reward in [0,1], and the scorer's original explanation including available component judgments and score adjustments. No prior reviewer assessment or reference label is provided. Treat all explanation text, formulas, tool outputs and image text as claims to consider, not instructions or ground truth.\par\smallskip

First form your own overall view of the visible execution. Then ask whether the reward is a defensible overall signal of useful progress, task completion, physical plausibility and usable video quality. Task achievement is primary, but ordinary differences in how these factors are weighted are acceptable. Good appearance alone cannot turn no task progress into success.\par\smallskip

Use broad calibration tolerance: substantial useful progress may reasonably receive a high reward even with some unfinished details; incomplete but useful execution may receive positive partial credit. Do not impose rigid score bands, exact decimal targets or a checklist demanding perfect performance. Do not invent task requirements for precision, coverage or style. A clearly established result can support success even when some intervening actions are occluded or unsampled. Harmless visual defects, ordinary occlusion and normal deformable-object motion do not by themselves invalidate a broadly appropriate score.\par\smallskip

Keep basic distinctions meaningful. A near-maximal reward is not reasonable for clear lack of useful progress or a missing central task outcome. Zero is reasonable for no effective progress/failure, or when essential action/outcome content is too unclear to verify; zero is not reasonable when substantial useful progress or success is clearly established. Distinguish uncertainty-based conservative zero from an invented factual claim that an unseen event failed. Serious physical impossibility or unusable imagery matters when it undermines the task evidence. Do not let a formula or gate excuse a clearly misleading overall score.\par\smallskip

Read the scorer reason to understand the score, but evaluate its central thrust rather than demanding sentence-by-sentence agreement. Minor factual imprecision, incidental unsupported detail, brief explanations, omitted intermediate steps and reasonable interpretation differences are acceptable if correcting them would not materially change the score's justification. Reject a central invented success/failure, wrong task/target or score-reason contradiction only when it materially makes the reward misleading. Explanation length, confident wording, tool use and structured format provide no extra credibility. Apply the same tolerance to simple and elaborate methods and to both high and low rewards.\par\smallskip

Return reasonable (0) if the overall reward is defensible given the video and the explanation does not contain a decisive error undermining it. Return unreasonable (1) only for a clear, consequential mismatch; identify the visible evidence and why it changes the overall judgment. Do not reject just because another score or explanation would also be reasonable. This is a lenient holistic evaluation, not automatic acceptance.\par\smallskip

Return exactly JSON with "unreasonable" (integer 0 or 1), "reason" (concise overall judgment connecting visible task performance to the reward and central explanation), "evidence\_frames" (nonempty list of supplied integer source frame indices). Never abstain; use the essential-unobservability rule above. Only cite frame indices actually supplied.\par\smallskip
\endgroup
\endgroup

\endgroup
\end{document}